\PassOptionsToPackage{table}{xcolor}
\PassOptionsToPackage{pagebackref,breaklinks=true,colorlinks,citecolor=blue,urlcolor=blue,linkcolor=blue,bookmarks=false}{hyperref}
\PassOptionsToPackage{noabbrev,nameinlink,capitalize}{cleveref}

\documentclass[onecolumn, numbers]{april_aigc}

\usepackage[utf8]{inputenc} 
\usepackage{url}           
\usepackage{amsfonts}       
\usepackage{amssymb}        
\usepackage{amsmath}       
\usepackage{nicefrac}      
\usepackage{microtype}      
\usepackage{xspace}         
\usepackage{fix-cm}
\usepackage[T1]{fontenc}

\usepackage{booktabs}      
\usepackage{wrapfig}       
\usepackage{multicol}       
\usepackage{multirow}       
\usepackage{makecell}       
\usepackage{tabularx}       
\usepackage{adjustbox}      
\usepackage{colortbl}       
\usepackage{pifont}         
\usepackage{enumitem}
\usepackage[normalem]{ulem}

\usepackage{xcolor}
\definecolor{linkcolor}{named}{aprilblue}
\definecolor{urlcolor}{RGB}{255,105,180}
\definecolor{citecolor}{RGB}{66,168,235}
\definecolor{lightgray}{rgb}{0.8, 0.8, 0.8}
\definecolor{darkgreen}{rgb}{0.00, 0.81, 0.78}

\definecolor{gray_tab}{RGB}{220, 220, 220}
\definecolor{blue_tab}{RGB}{227, 240, 251}
\definecolor{oran_tab}{RGB}{252, 242, 237}
\definecolor{whit_tab}{RGB}{255, 255, 255}
\definecolor{green_code}{RGB}{55, 126, 34}

\usepackage{algorithm}
\usepackage{algorithmic}
\usepackage{listings}
\usepackage{etoolbox}

\makeatletter
\AfterEndEnvironment{algorithm}{\let\@algcomment\relax}
\AtEndEnvironment{algorithm}{\kern2pt\hrule\relax\vskip3pt\@algcomment}
\let\@algcomment\relax
\newcommand\algcomment[1]{\def\@algcomment{\footnotesize#1}}
\renewcommand\fs@ruled{\def\@fs@cfont{\bfseries}\let\@fs@capt\floatc@ruled
  \def\@fs@pre{\hrule height.8pt depth0pt \kern2pt}%
  \def\@fs@post{}%
  \def\@fs@mid{\kern2pt\hrule\kern2pt}%
  \let\@fs@iftopcapt\iftrue}
\makeatother

\usepackage[pagebackref,breaklinks=true,colorlinks,citecolor=blue,urlcolor=blue,linkcolor=blue,bookmarks=false]{hyperref}
\AtEndPreamble{
    \usepackage[capitalize]{cleveref}
    \crefname{section}{Sec.}{Secs.}
    \Crefname{section}{Section}{Sections}
    \crefname{table}{Tab.}{Tabs.}
    \Crefname{table}{Table}{Tables}
    \crefname{equation}{Eq.}{Eqs.}
    \Crefname{equation}{Equation}{Equations}
    \crefname{figure}{Fig.}{Figs.}
    \Crefname{figure}{Figure}{Figures}
}
\hypersetup{colorlinks=true,linkcolor=linkcolor,urlcolor=urlcolor,citecolor=citecolor}

\usepackage{caption}
\DeclareCaptionFormat{custom}{{\color{aprilblue}\sffamily\textbf{#1 #2}} #3}
\titleformat*{\section}{\color{aprilblue}\Large\sffamily\bfseries}
\titleformat*{\subsection}{\color{aprilblue}\large\sffamily\bfseries}
\titleformat*{\subsubsection}{\color{aprilblue}\normalsize\sffamily\bfseries}

\usepackage{fancyhdr}

\newif\ifshowlogo
\showlogotrue   
\showlogofalse

\newcommand{\insertlogo}{%
  \ifshowlogo
    \IfFileExists{assets/april_logo1.png}%
    {\includegraphics[height=0.68cm]{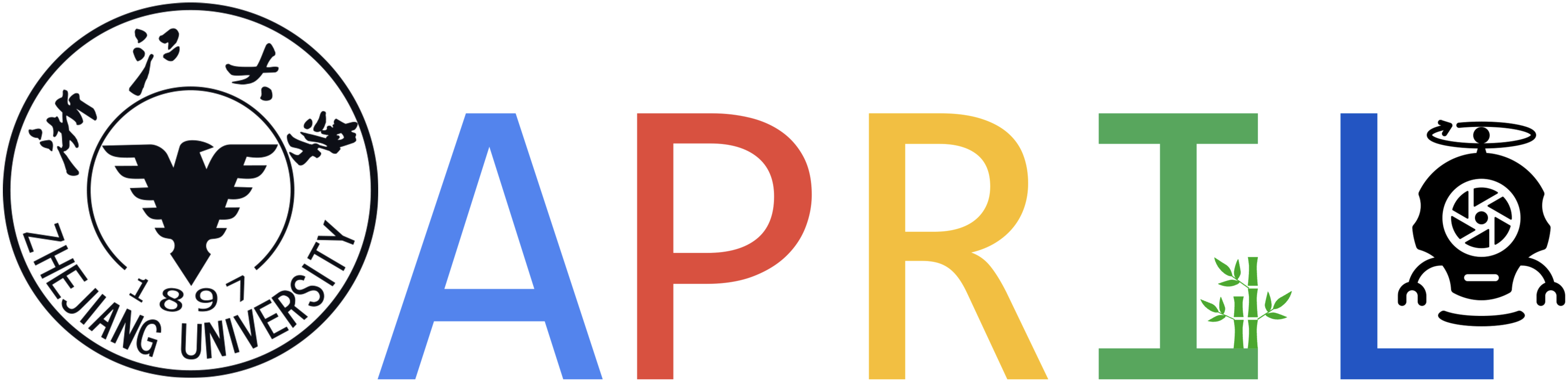}}
    {}%
  \fi
}

\newif\ifshowtoc
\showtocfalse

\renewcommand{\title}[1]{\def\titlelist{{\fontsize{20pt}{28pt}\selectfont\sffamily\bfseries #1}}}
\title{Multi-Dimensional Knowledge Profiling with Large-Scale Literature Database and Hierarchical Retrieval}

\author[1]{Zhucun Xue}
\author[1,\raisebox{-0.2em}{\includegraphics[height=0.85em]{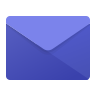}}]{Jiangning Zhang}
\author[1]{Juntao Jiang}
\author[1]{Jinzhuo Liu}
\author[1]{Haoyang He}
\author[2]{Teng Hu}
\author[3]{Xiaobin Hu}
\author[1,\raisebox{-0.2em}{\includegraphics[height=0.85em]{assets/icons8-email-96.png}}]{Yong Liu}
\author[3]{Shuicheng Yan}

\affiliation[1]{Zhejiang University, APRIL Lab}
\affiliation[2]{Shanghai Jiao Tong University}
\affiliation[3]{National University of Singapore}

\abstract{
The rapid expansion of research across machine learning, vision, and language has produced a volume of publications that is increasingly difficult to synthesize. Traditional bibliometric tools rely mainly on metadata and offer limited visibility into the semantic content of papers, making it hard to track how research themes evolve over time or how different areas influence one another.
To obtain a clearer picture of recent developments, we compile a unified corpus of more than 100,000 papers from 22 major conferences between 2020 and 2025 and construct a multidimensional profiling pipeline to organize and analyze their textual content. By combining topic clustering, LLM-assisted parsing, and structured retrieval, we derive a comprehensive representation of research activity that supports the study of topic lifecycles, methodological transitions, dataset and model usage patterns, and institutional research directions.
Our analysis highlights several notable shifts, including the growth of safety, multimodal reasoning, and agent-oriented studies, as well as the gradual stabilization of areas such as neural machine translation and graph-based methods. These findings provide an evidence-based view of how AI research is evolving and offer a resource for understanding broader trends and identifying emerging directions. 
}

\coverdate{\today}
\covercorrespondence{\email{12432038@zju.edu.cn}}
\coversourcecode{https://github.com/xzc-zju/psl}
\coverdataset{https://huggingface.co/datasets/APRIL-AIGC/psl}
\coverdatasetmodelscope{https://modelscope.cn/datasets/APRIL6AIGC/psl}

\begin{document}

\maketitle
\thispagestyle{plain}

\ifshowtoc
    \clearpage

    \setcounter{tocdepth}{2} 
    
    \tableofcontents
    \vspace{1cm} 

    \clearpage
   
\fi

\section{Introduction} \label{sec:introduction}
The scale and diversity of contemporary AI research continue to grow at an extraordinary pace. Across computer vision, machine learning, natural language processing, and related areas, the past five years have seen rapid shifts in model architectures, training strategies, datasets, benchmarks, and application domains. For researchers, this expansion brings substantial challenges: it is increasingly difficult to situate individual works within broader developments, to track how research themes evolve, or to identify areas that are emerging, stabilizing, or declining.

Existing tools only partially address this need. Traditional bibliometric methods, which built on metadata, co-citation networks, and keyword statistics~\cite{zupic2015bibliometric,donthu2021conduct,grootendorst2022bertopic}, provide high-level overviews but capture limited semantic information and treat topics as largely static entities. As a result, they offer only coarse insight into methodological transitions, cross-domain influences, or the finer-grained structure of research problems. Meanwhile, recent systems that incorporate large language models (LLMs)~\cite{brown2020language,achiam2023gpt} demonstrate improved semantic analysis, supporting tasks such as retrieval-augmented question answering~\cite{gao2023retrieval} and automated survey generation~\cite{wang2024autosurvey}. However, these tools are typically designed for short-range retrieval, single papers, or narrow tasks, and they do not provide a coherent, longitudinal view of large scientific corpora.

\begin{wrapfigure}{r}{0.5\textwidth}
    \centering 
    \includegraphics[width=1.0\linewidth]{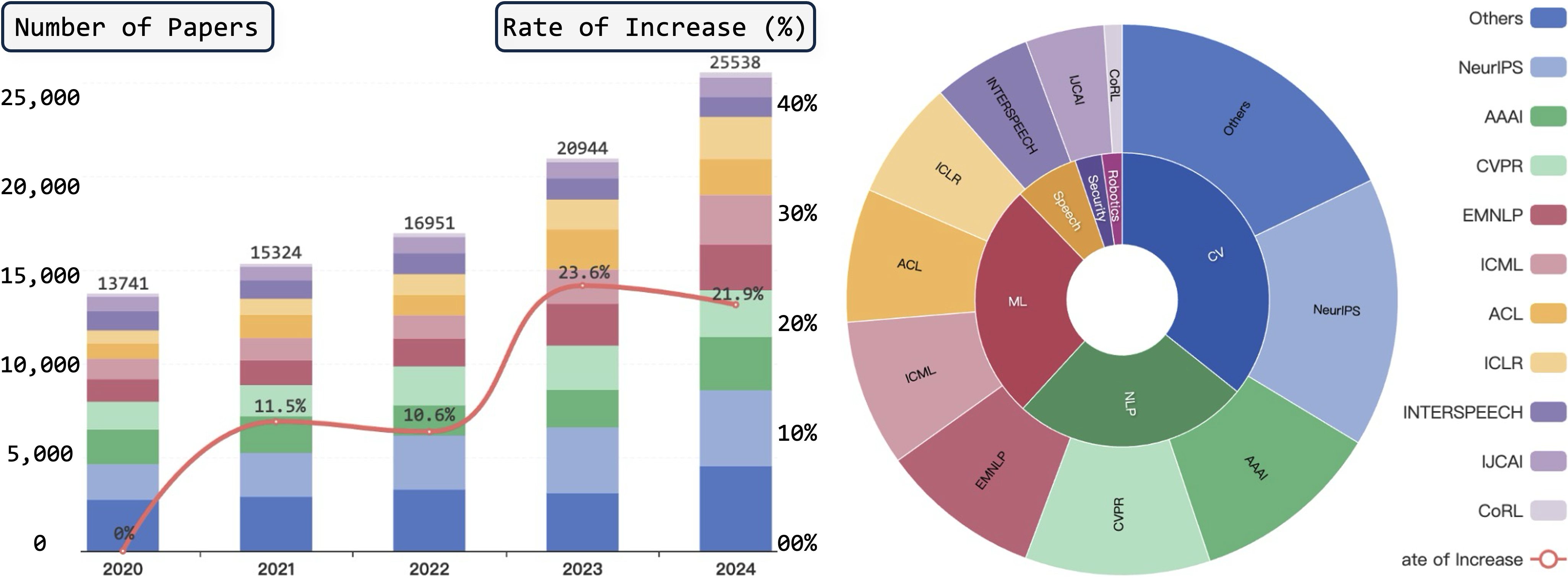} 
    \caption{ Number of papers published and topic statistics across 22 conferences from 2020 to 2025. } 
    \label{fig:paper_num} 
    \vspace{-0.5em} 
\end{wrapfigure}

These gaps point to the need for a unified way to organize, summarize, and interpret the rapidly expanding body of AI literature. In this work, we construct a large-scale profiling pipeline aimed at characterizing the recent development of AI research. Using more than 100,000 papers from 22 major conferences published between 2020 and 2025 (\cref{fig:paper_num}), we combine text clustering, LLM-assisted semantic parsing, and lightweight retrieval techniques to form a structured representation of research problems, methods, datasets, and topical dynamics. 
Rather than emphasizing algorithmic novelty, our focus is on creating a coherent analytic framework that enables researchers to explore and reason about the field at multiple levels of granularity.

Our study provides two complementary benefits. First, we derive a high-resolution view of topic lifecycles, dataset and model adoption patterns, and methodological transitions across areas such as vision, multimodal learning, foundation models, and generative modeling. Second, by incorporating structured retrieval and semantic filtering, we enable grounded, evidence-based queries that support practical research tasks, such as surveying subfields, tracing the evolution of techniques, or identifying emerging directions.

Through this analysis, we highlight several notable shifts in the AI landscape, including the consolidation of previously fast-moving areas, the rise of multimodal and agent-oriented research, and clear transitions in compute usage and model scaling practices. We expect the resulting knowledge database to serve as a resource for understanding broad trends, informing future meta-analyses, and supporting data-driven research planning.

Our work makes the following contributions:

\begin{itemize}
\item We construct a large-scale profiling pipeline over more than 100{,}000 papers from 22 major conferences, enabling structured analysis of semantic topics, methods, datasets, and research trajectories.
\item We integrate clustering-based topic organization with LLM-assisted parsing and retrieval, producing a structured and interpretable representation of the AI research landscape that complements traditional bibliometrics.
\item We conduct comprehensive empirical analyses, revealing trends in topic evolution, emerging subfields, methodological transitions, dataset and model dynamics, and institutional research patterns across the broader AI community.
\end{itemize}

Overall, these results provide an evidence-based view of how modern AI research is evolving and offer a foundation for transparent, large-scale, and semantically grounded scientometric analysis.

\section{Related Work} \label{sec:related_work}

\subsection{Bibliometrics \& Scientific Trend Analysis}
Content mining and trend analysis of scientific literature are important topics in scientometrics and information science. Traditional bibliometric approaches rely on metadata, extracting abstracts, authors, journals, keywords, and citations\cite{zupic2015bibliometric,donthu2021conduct}, and applying co-citation and co-word analyses\cite{van2010software,chen2006citespace,aria2017bibliometrix} to reveal the structure and evolution of research fields. However, these methods mainly rely on surface features and are limited in capturing the latent semantic structures within documents. To address this, topic modeling and unsupervised clustering have been widely used\cite{blei2012probabilistic}, with LDA inferring topic distributions from word co-occurrences\cite{blei2003latent,alsumait2009topic,chang2009reading,chuang2013topic}. More recently, embedding-based clustering methods, such as BERTopic\cite{grootendorst2022bertopic,reimers2019sentence} combined with BERT\cite{devlin2019bert}, and applications of large language models\cite{kostikova2025lllms,diaz2025k,ccelikten2025topic,lam2024concept} have enhanced the discovery of latent topics. Nevertheless, existing methods still struggle to systematically represent the dynamic evolution of knowledge\cite{si2024can,scherbakov2025emergence}, and a unified framework for multidimensional knowledge profiling remains lacking.

\subsection{LLMs for Document Understanding}

Trained on massive general-domain corpora, LLMs have shown strong abilities in semantic understanding and text generation \cite{brown2020language}, giving rise to series of GPT \cite{achiam2023gpt} and other state-of-the-art models \cite{adakd,comanici2025gemini,touvron2023llama,yang2025qwen3}. To support scientific applications, researchers have explored LLMs in research workflows. For example, PaperQA \cite{lala2023paperqa,skarlinski2024language,besrour2025squai} leverages retrieval-augmented generation (RAG) techniques \cite{adavideorag,gao2023retrieval,li2025towards,cheng2025survey} to enable document question answering, while multi-agent and reinforcement learning approaches \cite{nguyen2025ma,singh2025agentic,li2025towards,wu2025structure} further improve system accuracy. Another line of work focuses on automating literature review generation, including AutoSurvey \cite{wang2024autosurvey}, SurveyX\cite{liang2025surveyx}, SurveyForge \cite{liang2025surveyx,yan2025surveyforge}, and SurveyG \cite{nguye2025surveyg}, which implement pipelines for retrieval, filtering, organization, and writing. 
Despite these advances, existing LLM-based approaches are generally limited in scope: they focus on single tasks, rely on predefined corpora, and seldom provide a unified view of scientific knowledge evolution across multiple dimensions, including method evolution, task evolution, dataset adoption, and compute trends. Moreover, few approaches systematically integrate temporal information, semantic structure, and cross-conference knowledge, limiting their utility for large-scale trend analysis and evidence-based research decision making.

In this work, we propose a multidimensional profiling framework that leverages LLMs to extract and organize semantic information from a large corpus of scientific publications. By combining embeddings, topic clustering, and hierarchical retrieval, the framework produces a dynamic, hierarchical view of knowledge evolution, enabling trend analysis, cross-domain comparisons, and fine-grained topic investigation, and supporting data-driven exploration and research planning.

\section{Knowledge Profiling Framework} \label{sec:method}

\begin{figure}[tp]
  \centering
  \vspace{-1.5em} 
  \includegraphics[width=0.95\textwidth]{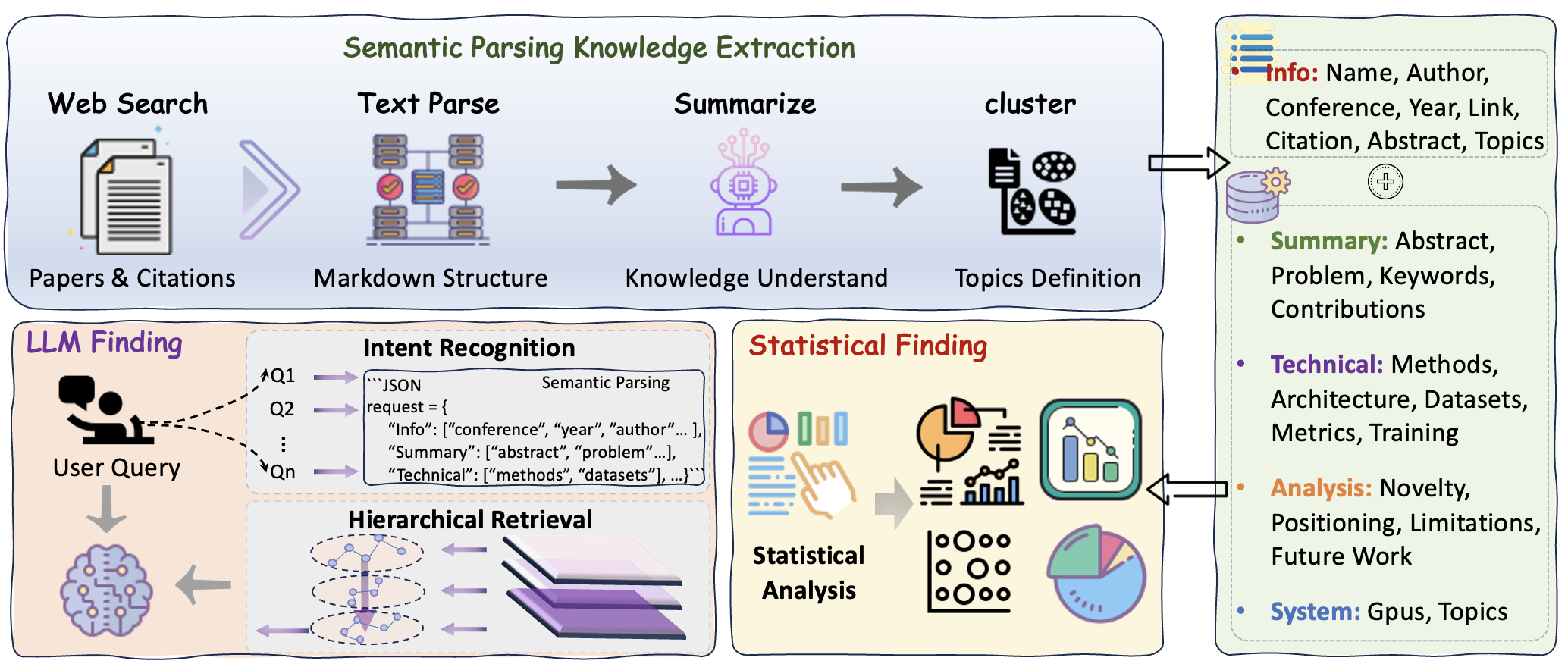} 
  \caption{ LLM-Driven Multi-Dimensional Knowledge Profiling Framework with Large-Scale Database and Hierarchical Retrieval. } 
  \label{fig:method} 
  \vspace{-0.5em} 
\end{figure} 

\begin{table*}[!htp]
\centering
\caption{Detailed annual paper distribution for the 22 major venues in $ResearchDB$ from 2020 to 2025. }

\resizebox{\textwidth}{!}{%
\begin{tabular}{lrrrrrrr}
\toprule
\textbf{Venue Group} & \textbf{2020} & \textbf{2021} & \textbf{2022} & \textbf{2023} & \textbf{2024} & \textbf{2025} & \textbf{Total} \\
\midrule
Core ML/Theory (NeurIPS, ICML, ICLR, UAI, COLT) & 264 & 4703 & 5587 & 7317 & 9596 & 7269 & 34736 \\
Computer Vision (CVPR, ICCV, ECCV) & 1358 & 3272 & 3715 & 4509 & 4868 & 2871 & 20593 \\
NLP/Speech (ACL, EMNLP, NAACL, COLM, INTERSPEECH, IWSLT) & 3098 & 4138 & 4498 & 5580 & 6417 & 4730 & 28461 \\
General/Applied AI (AAAI, IJCAI, CoRL) & 2644 & 2837 & 2683 & 3071 & 4178 & 3239 & 18652 \\
Systems/Security (MLSYS, OSDI, NDSS, etc.) & 216 & 374 & 468 & 467 & 479 & 36 & 2040 \\
\midrule
\textbf{Overall Total (22 Venues)} & \textbf{7580} & \textbf{15324} & \textbf{16951} & \textbf{20944} & \textbf{25538} & \textbf{18145} & \textbf{104482} \\
\bottomrule
\end{tabular}}
\label{tab:paper_nums}
\vspace{-0.5em}
\end{table*}

We collected papers accepted between 2020 and 2025 from 22 academic conferences, including CVPR, ECCV, ICCV, ICLR, NeurIPS, and ICML, along with the citation counts for each paper, and over 100,000 papers are included in total, as shown in \cref{tab:paper_nums}. Our main focus is on how to efficiently perform Multi-Dimensional Knowledge Profiling on this scientific data, as illustrated in \cref{fig:method}.

\subsection{Semantic Parsing \& Knowledge Extraction}

To systematically analyze the rapidly growing body of AI research literature, we propose a multi-stage knowledge extraction framework that combines automated content mining with LLM-driven semantic analysis. First, considering the difficulties in structured output and efficiency when directly using LLMs to analyze PDF files, we employ minerU \cite{wang2024mineru} to convert the collected PDFs into structured Markdown files. Subsequently, based on the Markdown file corresponding to each paper, we utilize Deepseek-R1-32B \cite{guo2025deepseek} to conduct multi-dimensional analysis of the papers, including meta information, abstract, research questions, methods, datasets, evaluation metrics, model architecture, contributions, limitations, and future directions. 

It is worth noting that in multi-dimensional content parsing, the topic information of the papers is crucial. However, when LLMs process large scale structured data, on the one hand, due to the hallucination problem, LLMs struggle to stably output topic classifications, and on the other hand, the processing speed is limited. To address this, we adopt a clustering method similar to BERTopic\cite{grootendorst2022bertopic} for topic clustering of papers. The specific process is as follows: first, use text encoder to process the title and abstract of each paper, then perform dimensionality reduction via UMAP\cite{mcinnes2018umap}, and then use HDBSCAN\cite{mcinnes2017hdbscan} for clustering, resulting in more than 300 topic categories distinguished based on semantic similarity. Finally, we use ChatGPT-5 to generate topic summaries for each cluster, extract core topic names, and construct a hierarchical logical relationship of topics, forming a multi-dimensional knowledge database, $ResearchDB$, which can be used for subsequent analysis. 
The ResearchDB schema captures five main dimensions (Summary, Technical, Analysis, System, Post-Processed) extracted from full paper texts using LLM-assisted parsing. \cref{tab:Schema} provides the detailed documentation for the key fields in our structured data records. Unlike traditional bibliometric methods, this framework relies on semantic understanding rather than mere keywords, enabling the discovery of emerging or previously underexplored research topics, and finally obtaining the following information:

\begin{table*}[h]
\centering
\scriptsize 
\caption{Revised documentation for the key structural fields in the ResearchDB schema, reflecting the specific output of the LLM-assisted parsing pipeline.}

\begin{tabularx}{\textwidth}{lcccX}
\toprule
\textbf{Field Name} & \textbf{Data Type} & \textbf{Source} & \textbf{Purpose/Description} \\
\midrule
\texttt{paper\_name/authors} & String/List & Metadata & Official title and author list. \\
\texttt{conference/year} & String/Int & Metadata & Publication venue and year. \\
\midrule

\texttt{keywords} & List[String] & LLM-Parsed & Essential technical terms and concepts identified in the full text. \\
\texttt{keywords\_description} & List[String] & LLM-Parsed & Essential technical terms and concepts identified in the full text. \\
\texttt{abstract\_ori} & String & LLM-Parsed & Detailed summary covering the paper's context, method, and final results. \\
\texttt{abstract\_summary} & String & LLM-Parsed & Detailed summary covering the paper's context, method, and final results. \\
\texttt{problem\_statement} & String & LLM-Parsed & The key issue the paper attempts to solve. \\
\texttt{contributions} & List[String] & LLM-Parsed & Explicit statements of the paper's primary research contributions. \\
\texttt{methods} & String & LLM-Parsed & Core algorithmic approach/methodology used (e.g., PAGE, SGD). \\
\texttt{architecture} & String & LLM-Parsed & Specific model architectures referenced or introduced (e.g., LeNet, ResNet). \\
\texttt{loss\_function} & String & LLM-Parsed & The specific loss function used for training (if mentioned). \\
\texttt{training\_setup} & String & LLM-Parsed & Details on optimization, batch size, and general training configuration. \\
\texttt{datasets} & List[String] & LLM-Parsed & List of datasets used for experimental validation. \\
\texttt{metrics} & List[String] & LLM-Parsed & Key performance indicators reported (e.g., Test Accuracy, Training Loss). \\
\texttt{gpu\_info} & String & LLM-Parsed & Specific hardware information referenced, used for resource analysis. \\
\texttt{limitations} & List[String] & LLM-Parsed & Identified constraints, weaknesses, or scope limitations of the proposed work. \\
\texttt{field\_positioning} & String & LLM-Parsed & Categorization of the paper's innovation type. \\

\midrule
\texttt{topic\_name/ID} & String/Int & Post-Processed & Hierarchical topic label assigned via clustering and LLM refinement. \\
\bottomrule
\end{tabularx}
\vspace{-1.5em}
\label{tab:Schema}
\end{table*}

\begin{itemize}
  \item \textbf{Info}: Extracts metadata including title, authors, conference, year, citations, topics, abstract, paper link, enabling macro-level analyses of research output, impact, and institutional distribution.
  \item \textbf{Summary}: Captures the core content, including research problem, keywords, abstract, and contributions, supporting the identification of emerging topics and cross-task trends.
  \item \textbf{Technical}: Extracts model architectures, datasets, evaluation metrics, and training strategies, allowing the study of methodological evolution and dataset usage patterns.
  \item \textbf{Analysis}: Summarizes novelty, positioning, limitations, and future work, facilitating the discovery of research gaps, bottlenecks, and potential directions.
  \item \textbf{System}: Records GPU, computational resources, enabling insights into the resource demands of different approaches and trends in large-scale model development.
\end{itemize}

\subsection{Intent-Driven Knowledge Retrieval}

\begin{wrapfigure}{r}{0.50\textwidth}
    \centering
    \vspace{-2.5em}
    \includegraphics[width=1.0\linewidth]{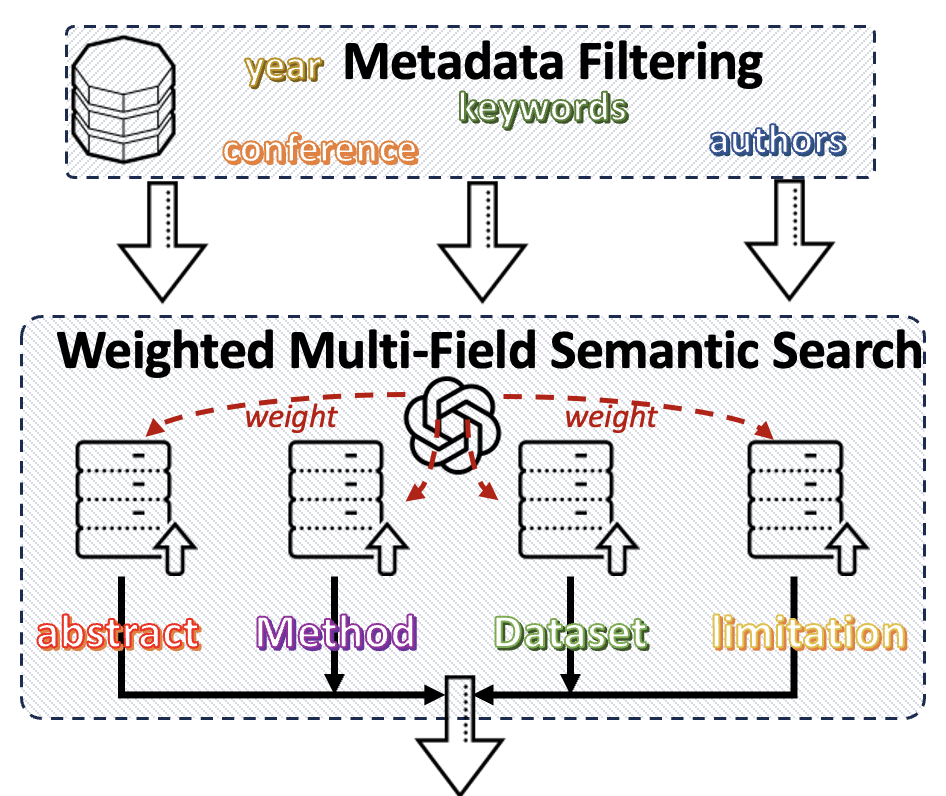}
    \caption{The framework of Hierarchical Retrieval.
    }
    \vspace{-1.5em}
    \label{fig:retial}
\end{wrapfigure}

Large language models have shown potential for literature retrieval and summarization, but direct application in research workflows faces key limitations: hallucination, omission or misinterpretation of complex information, and poor grounding in evidence, particularly over large, multi-year corpora. To address these challenges, we develop an intent-driven hierarchical retrieval pipeline over our structured $ResearchDB$, combining metadata filtering with weighted multi-field semantic search to provide reliable, evidence-based input to language models.

\noindent\textbf{Intent Recognition} 
To handle complex queries $Q$, we first decompose $Q$ into simpler sub-questions $\{Q_i\}$ that isolate distinct information needs. Each sub-question is then processed using semantic parsing to generate structured retrieval instructions in JSON format, specifying relevant keywords, entities, and content types for subsequent retrieval. This decomposition reduces ambiguity, ensures comprehensive coverage, and provides clear guidance for the hierarchical retrieval stage.

\noindent\textbf{Hierarchical Retrieval} 
Each sub-question $Q_i$ is addressed through a two-level retrieval strategy, as shown in\cref{fig:retial}.

\begin{enumerate}
    \item \textbf{Metadata Filtering.} We filter $ResearchDB$ using metadata attributes such as conference, year, authors, and keywords to obtain a candidate subset. Among them, keywords use fuzzy matching, and other fields use exact matching.
    \begin{equation}
    D_{\text{filtered}} = \text{Filter}(ResearchDB, \text{metadata}).
    \end{equation}

    \item \textbf{Weighted Multi-Field Semantic Search.} 

For the document subset $D_{\text{filtered}}$ obtained via metadata filtering, we perform a subsequent intent-driven semantic search. Unlike traditional methods with static weights, our framework employs an adaptive weighting mechanism where the field weights $w_j$ are dynamically assigned by the LLM-based intent recognition module according to the query's emphasis. For instance, if a sub-question $Q_i$ focuses on specific technical implementations, the module allocates significantly higher priority to the 'methods' field (e.g., $w_{\text{methods}} = 0.8$) relative to the 'abstract' (e.g., $w_{\text{abstract}} = 0.2$). These weights are then applied to the semantic similarity scores $s_{ij}$ calculated between the query and each core structural field—including abstract, methods, datasets, and limitations. The final weighted relevance score $S_i$ is computed as:

    \begin{equation}
    S_i = \sum_{j=1}^{n} w_j \cdot s_{ij}, \quad \text{with} \quad \sum_{j=1}^{n} w_j = 1.
    \end{equation}
    Top-$k$ documents based on $S_i$ are retrieved:
    \begin{equation}
     R_i = \text{TopK}(D_{\text{filtered}}, S_i).
    \end{equation}
\end{enumerate}

\noindent The retrieved passages are compiled and passed to the language model to generate a coherent, evidence-grounded response. Weighting across fields allows the system to prioritize more informative content while retaining flexibility for diverse query types.

\section{Scientific Insights and Findings} \label{sec:exp}

\begin{figure}[tp]
    \centering
    \includegraphics[width=0.90\linewidth]{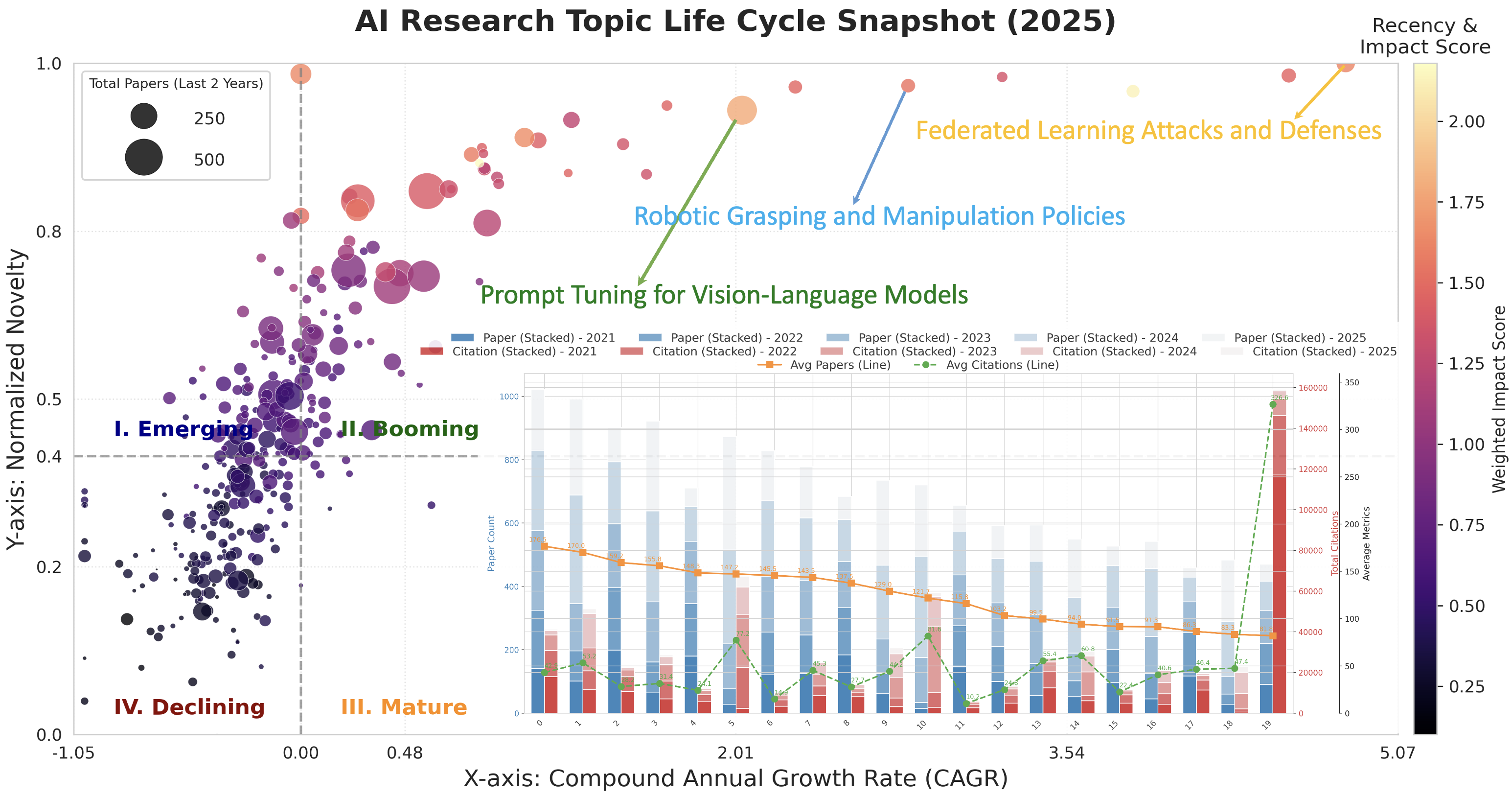}
    \vspace{-1em}
    \caption{Four-quadrant topic lifecycle evolution (quadrants I-IV: Emerging, Booming, Mature, Declining). The inset (bottom left) shows the most prolific topic by paper count from 2020-2025.
    }
    \label{fig:topic}
    \vspace{-1.5em}
\end{figure}
Leveraging our LLM-driven multidimensional profiling framework, we analyze more than 100K papers accepted by 20+ premier AI conferences from 2020 to 2025. This section presents a series of quantitative and semantic analyses across multiple dimensions of the research ecosystem, including topic evolution, compute and model scale, dataset dynamics, and institution-level research specialization. These results together reveal both the macro-level shifts and the fine-grained emerging directions shaping modern AI research.

\subsection{Topic Growth and Transitions}

To characterize the long-term evolution of AI research topics, we analyzed topic dynamics from 2020 to 2025 using LLM-based semantic clustering. We construct a four-quadrant topic lifecycle model, where the horizontal axis represents the Compound Annual Growth Rate (CAGR) of papers over the past two years, reflecting topic popularity trends:

\begin{equation}
\text{CAGR}_t = \left(\frac{N_t}{N_{t-2}}\right)^{\frac{1}{2}} - 1, \quad t = 2025.
\end{equation}

\noindent The vertical axis represents the normalized mean publication year $\overline{Y}$ of papers, capturing research novelty:

\begin{equation}
\overline{Y}_i = \frac{1}{N_i} \sum_{p \in \text{topic } i} Y_p.
\end{equation}

\noindent The bubble size indicates the total number of papers in the past two years $N_i$, and the color encodes a weighted impact metric combining citations $C_i$ and topic count $T_i$:

\begin{equation}
\text{W}_i = \frac{\alpha C_i + \beta T_i}{\max_j (\alpha C_j + \beta T_j)}, \quad \alpha + \beta = 1.
\end{equation}
\noindent $\alpha$ and $\beta$ control the relative weights of citations and paper count, with $\alpha = 0.6$ in our experiments.
 
By analyzing the distribution of various topics in the four quadrants, as shown in \cref{fig:topic}, we observe that AI research is accelerating its evolution from being driven by early "model scale" and "basic perception tasks" to a paradigm driven by "safety and controllability, multimodal cognition, and intelligent agent systems". To further verify the overall trend of the life cycle analysis, we counted the changes in the number of papers on various topics over the past five years, and aslo selected the Top-20 topics with the largest number of papers for time series analysis, as shown in the lower left in \cref{fig:topic}. 

The results show that the paper volume of LLM-related topics exhibits exponential growth, especially in reasoning, long-context modeling, and preference alignment, with each surpassing 200 papers per year by 2025. Multimodal integration tasks such as text-to-image editing, video diffusion, and vision-language QA maintain high activity but show slight stabilization after their 2023 to 2024 peaks. Meanwhile, classical areas including neural machine translation and GNN-based methods display relatively flat or declining trajectories, consistent with a maturing stage. Safety and robustness topics, though more recent, feature the steepest growth curves in both publication count and citation acceleration, marking the next frontier of attention.

Overall, from the dual perspectives of life cycle and quantity trends, the analysis reveals a clear paradigm transition in current research topics, from model architecture optimization toward the integration of perception, reasoning, and interaction capabilities. Future directions that deserve particular attention include:
\begin{itemize}
  \item Enhancing long-context and reasoning capabilities of foundation models through efficient memory and adaptive attention mechanisms;
  \item Strengthening safety, reliability, and alignment frameworks to ensure trustworthy human-AI collaboration;
  \item Advancing multimodal understanding and generation toward dynamic, interactive, and grounded cognition;
  \item Bridging efficiency and scalability with sustainable model deployment, including low-rank adaptation, quantization, and sparse computation strategies.
\end{itemize}
These directions exhibit both high recent growth rates and strong Weighted Impact values, suggesting sustained momentum in the evolving research landscape.

\subsection{Compute \& Model Scale Analysis}

\begin{figure*}[tp]
    \centering
    \includegraphics[width=0.87\linewidth]{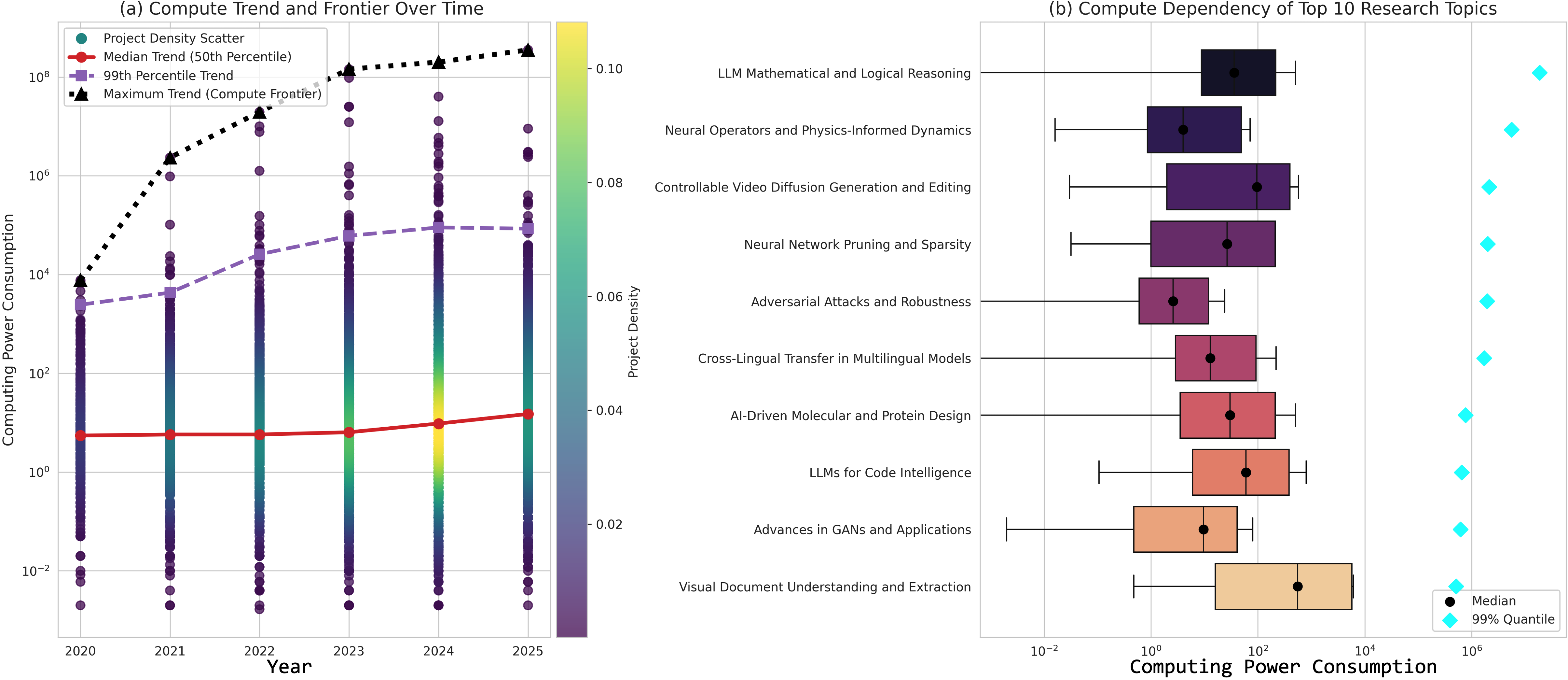}
    \vspace{-1em}
    \caption{
    Compute Resource Analysis. Left: Temporal evolution of compute demand; Right: Boxplot of the top 10 topics with highest compute requirements.
    }
    \label{fig:GPU}
    \vspace{-1em}
\end{figure*}   

With the rise of large models and the increase in complex tasks, the consumption of computing resources and the scale of model parameters have become important dimensions for measuring research trends. We analyzed the information on the use of computing resources in the statistical data and unified all computing power into A100 Equivalent Hours, as shown in \cref{fig:GPU}. The analysis results show that the consumption of training resources has maintained a significant upward trend, especially in related research such as generative models, multimodal models, and intelligent agents. This trend is consistent with the current pursuit of model scale and complexity in the AI field, reflecting the reliance of higher-performance models on huge computing power.

Our research found that topics requiring high computing power investment have the following characteristics.
\begin{itemize}
    \item \textbf{Processing complex data structures or large-scale data:} Research such as multimodality (combining images, text, speech, etc.) and large-scale pre-training involves massive amounts of training data with complex interrelationships between data, requiring more computing resources for feature extraction and model fitting.
    \item \textbf{Enlargement and deepening of model architectures: }Represented by large language models (LLMs) and large generative models, the number of parameters often reaches tens of billions or hundreds of billions, which directly leads to an exponential growth in the demand for computing resources during the training and inference phases.
    \item \textbf{Complex learning paradigms or training methods:} For example, research involving reinforcement learning (RL), adversarial learning (such as GANs), or research on intelligent agents that require a large number of interactions with simulated environments. These methods usually require multiple iterations, a large number of samples, or complex optimization processes, thereby greatly increasing computing power consumption.
\end{itemize}

Further analysis of the proportion of computing power and paper output across different topics reveals that the investment in computing resources shows an uneven distribution across various research directions:
\begin{itemize}
    \item \textbf{High-investment and high-output fields:} Concentrated in cutting-edge areas such as generative models and multimodality. Although they consume the majority of equivalent A100 hours, they also yield the highest number of papers and the most influential research results, forming a positive cycle where computing power drives innovation.
    \item  \textbf{High-investment and medium-to-low-output fields:} Research in certain intelligent agent or specific application domains may require enormous computing power for experiments due to immature methods or the inherent difficulty of tasks (such as exploration, the complexity of environmental interactions, etc.). However, the short-term paper output is relatively mismatched with their computing power consumption, which may represent future potential directions or bottlenecks in technological breakthroughs.
    \item  \textbf{Low-investment and high-efficiency fields:} Areas such as model compression, efficiency optimization, and interpretability focus on algorithms and efficiency rather than the absolute scale of models. These studies achieve a relatively high number of paper outputs with relatively low computing power investment, and are of great value for the promotion and practical application of existing models.
\end{itemize}

Overall, AI research is in a phase where its reliance on computing power continues to rise, mainly driven by large-scale models and complex tasks. Generative models, multimodality, and intelligent agents are the fields with the most intensive investment in computing power, indicating future research hotspots. At the same time, we have also observed that the academic community has achieved significant results in fields such as efficiency optimization and model simplification, which provide important support and alternative paths for promoting the development of AI in environments with limited resources.

\begin{figure}[tp]
  \centering
  \begin{subfigure}{0.48\textwidth}
    \centering
    \includegraphics[width=1.0\linewidth]{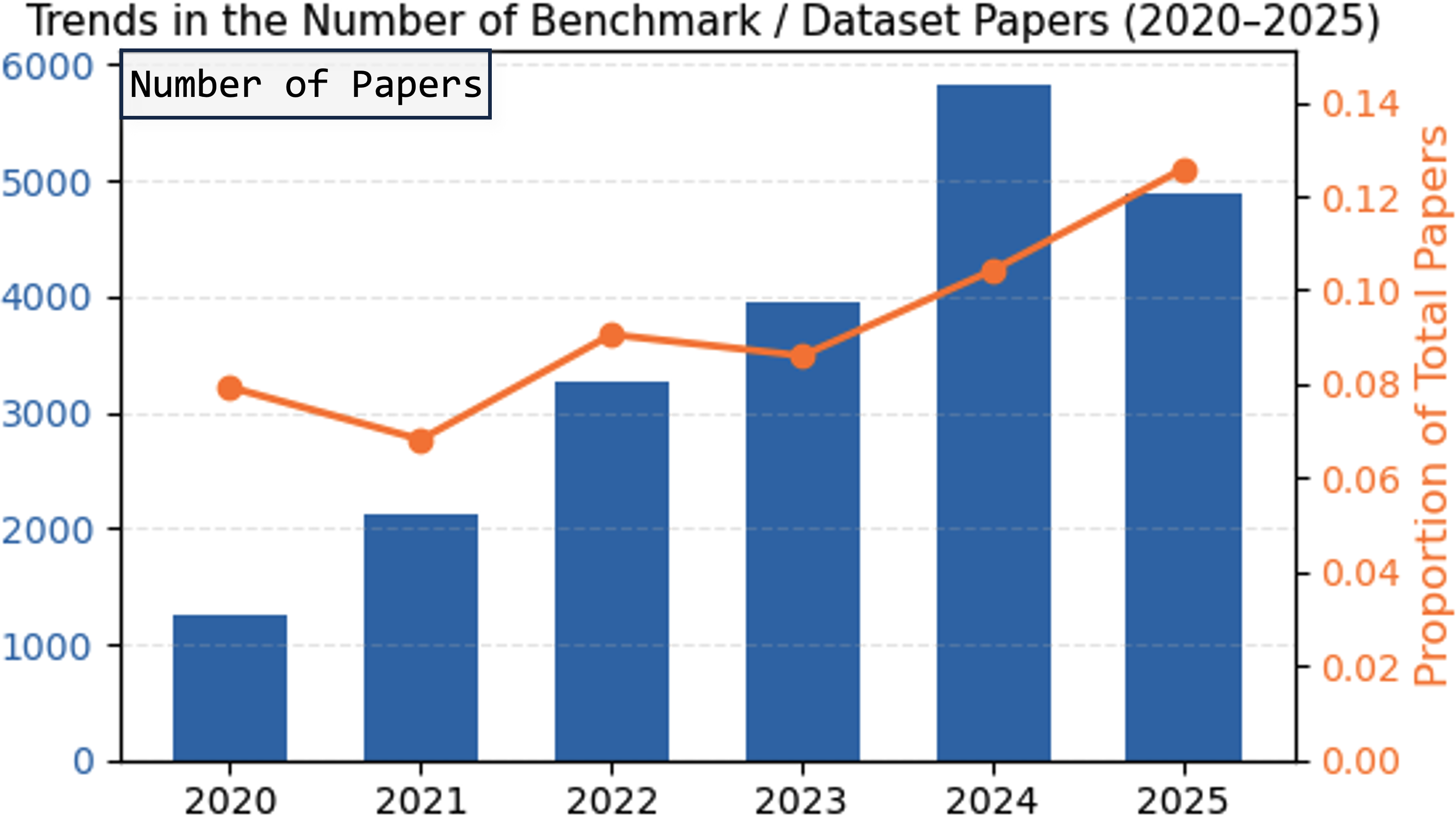}
     \vspace{-0.5em}
    \caption{
    Dataset and Benchmark Trends (2020-2025). The bar chart shows annual paper counts for each dataset type, and the line chart shows their proportion of total publications.
    }
    \label{fig:trend}
  \end{subfigure}
  \hfill 
  \begin{subfigure}{0.48\textwidth}
    \centering
      \includegraphics[width=1.0\linewidth]{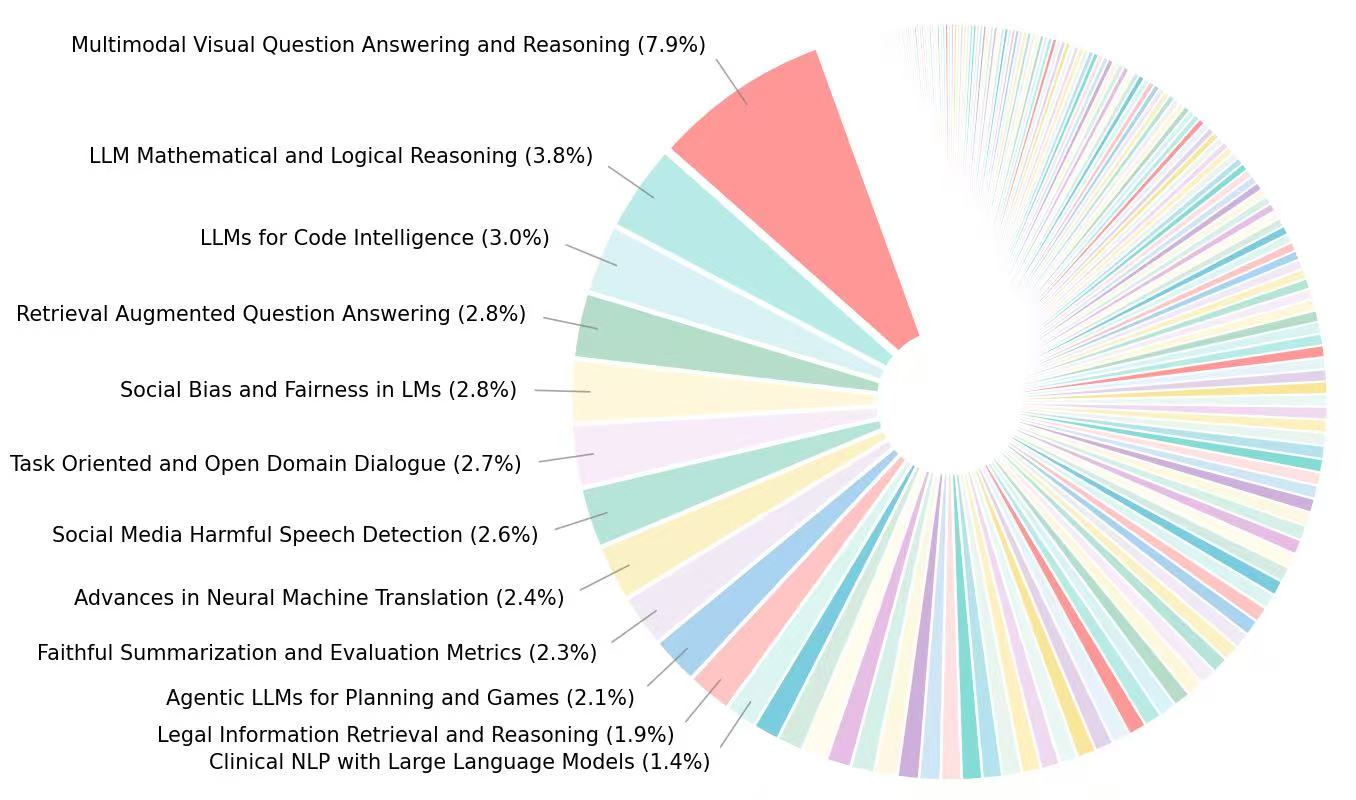}
    \caption{
   Topic distribution in Dataset/Benchmark papers. The most frequent topics are labeled.
    }
    \label{fig:pie}
  \end{subfigure}
  \caption{The Analysis from the Dataset Landscape.} 
  \label{fig:twofigs}
\end{figure}

\subsection{Dataset Landscape}

Benchmark papers lay a crucial foundation for the standardization and comparability of technological progress by providing unified evaluation criteria and datasets. Statistics from the sample set show that the proportion of Benchmark papers is on a significant upward trend, with the growth rate accelerating particularly after 2022, as shown in \cref{fig:trend}. Their proportion rose rapidly from 10.07\% in 2022 to 18.16\% in 2025 (partial data), approaching one-fifth of the total number of papers. This explosive growth is mainly due to the rise of general artificial intelligence technologies represented by LLMs, as well as the academic community's emphasis on the standardized evaluation of non-performance indicators such as model fairness, ethics, and robustness.

Further in-depth analysis of the themes of Benchmark articles reveals the evaluation challenges faced by current research hotspots and cutting-edge technologies. As shown in the statistical chart of theme count and proportion (\cref{fig:pie}), research focuses are highly concentrated on LLMs, multimodality, and complex reasoning tasks. The top ten themes are almost entirely centered around the latest advancements and applications of LLMs, fully reflecting the evaluation-driven characteristics of contemporary scientific research.

\begin{figure}[tp]
  \centering
  \begin{subfigure}{0.48\textwidth}
     \centering
    \includegraphics[width=0.95\linewidth]{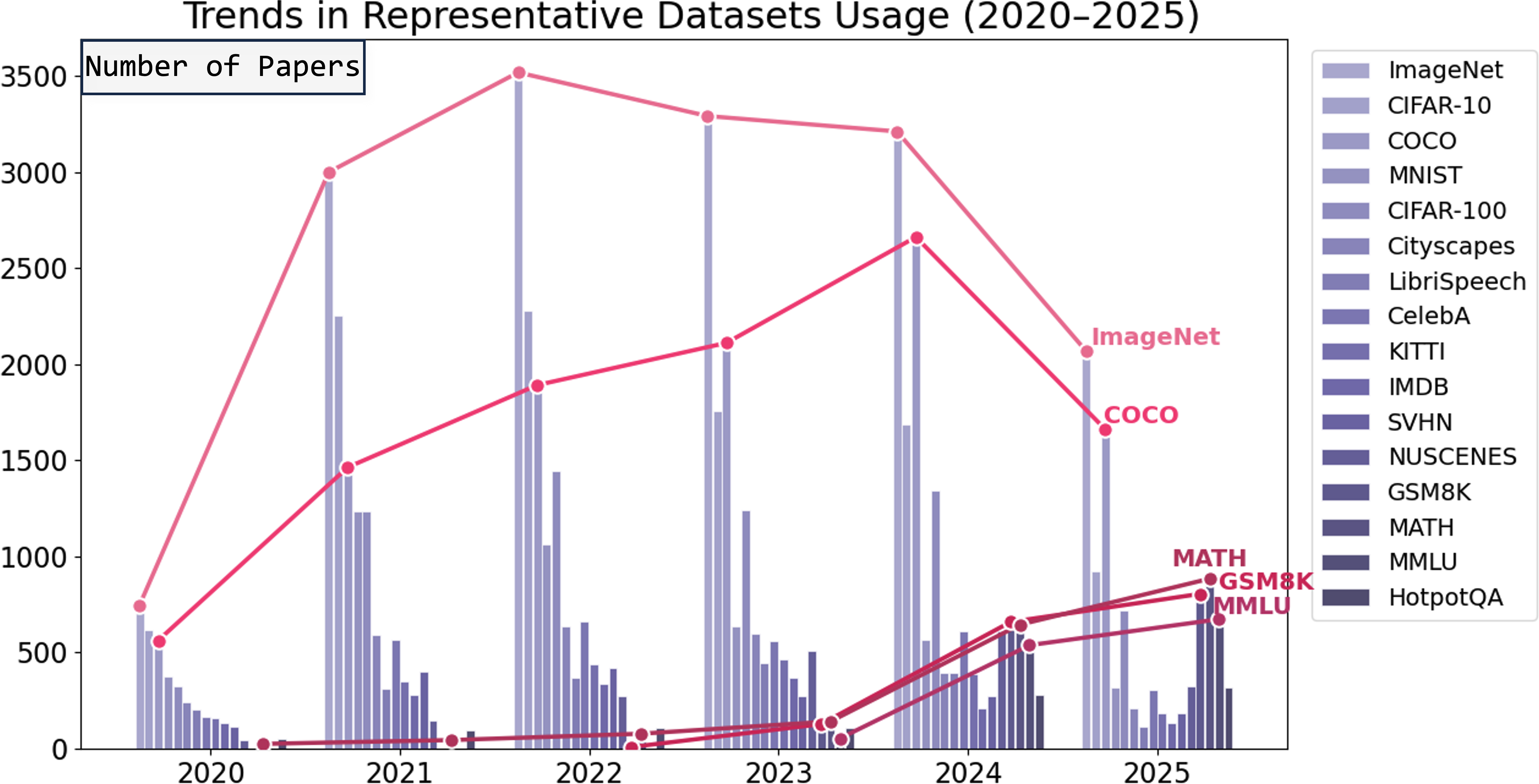}
    \vspace{-0.5em}
    \caption{
   Trends in Representative Dataset Usage (2020-2025). The grouped bar chart shows annual paper counts per dataset, with overlaid lines highlighting trends for MMLU, GSM8K, ImageNet, COCO, and MATH.
    }
    \label{fig:dataset_uasge}
  \end{subfigure}
  \hfill 
  \begin{subfigure}{0.48\textwidth}
    \centering
\includegraphics[width=0.95\linewidth]{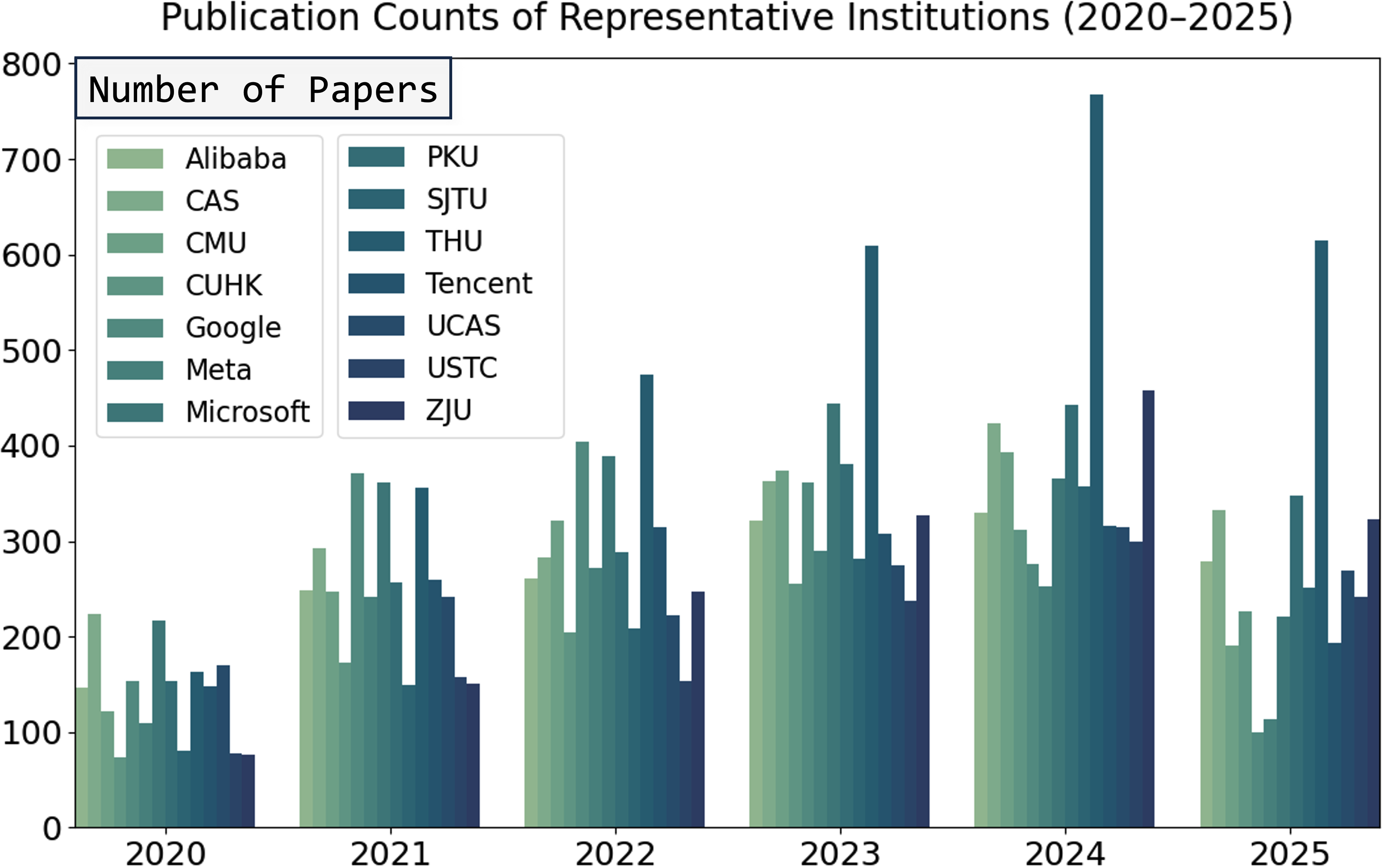}
    \vspace{-0.5em}
    \caption{
    Publication Counts of Leading Institutions (2020-2025). Shows the number of papers from representative academic and industrial institutions across 22 conferences.
    }
    \label{fig:ins_count}
  \end{subfigure}
  \caption{ Trends in Publication Counts for Dataset Usage and Leading Institutions from 2020 to 2025.} 
\end{figure}

Against the backdrop of this surge in benchmarks, the frequency of use of various datasets in scientific research papers has also shown rapid growth. This reflects the need for deeper evaluation in the era of LLMs and signifies that the paradigm for using datasets is no longer limited to single tasks but is evolving toward cross-domain and complex capability assessment, as shown in \cref{fig:dataset_uasge}

\begin{itemize}
    \item \textbf{Relative saturation of classic visual benchmarks and cross-modal applications. }Although ImageNet has long been the most widely used classic visual dataset, its peak appeared in 2022, indicating the relative saturation of traditional deep learning visual benchmarks. On the other hand, datasets such as COCO, which initially focused on traditional visual tasks like object detection and instance segmentation, have also been widely used for cross-modal capability evaluation since the rise of Multimodal LLMs. This shows that research on large models is promoting cross-task and multi-purpose use of datasets, utilizing the rich annotated resources of classic datasets to evaluate new comprehensive capabilities.
    \item \textbf{Explosive growth of high-level reasoning datasets.} As LLMs mature in basic language capabilities, research focus has shifted to complex cognitive and reasoning tasks. The number of papers using the MATH dataset, designed for evaluating mathematical reasoning and arithmetic abilities, as well as MMLU and GSM8K for high-level logical reasoning, has risen sharply since 2023. This trend strongly suggests that researchers are increasingly relying on these datasets to evaluate the performance of large models in complex reasoning tasks such as logic, common sense, and the integration of professional knowledge.
\end{itemize}

\subsection{Institution-Level Research Patterns}
\begin{wrapfigure}{r}{0.45\textwidth} 
    \centering
    \vspace{-2em}
    \includegraphics[width=1.0\linewidth]{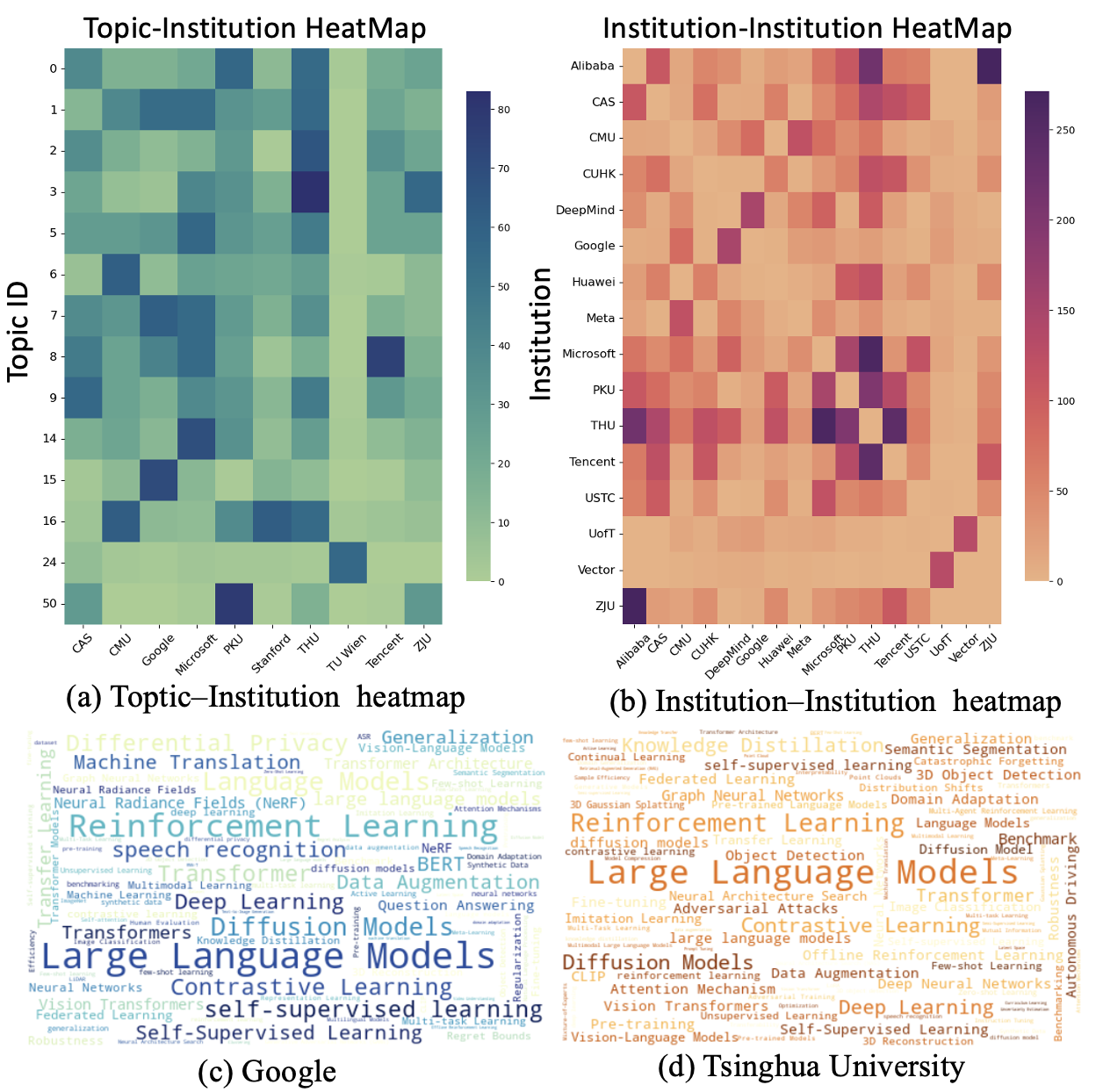}
     \vspace{-2em}
    \caption{(a)--(b) Top topic-institution and institution-institution heatmaps;
    (c)--(d) Research topic word clouds for Google and Tsinghua University.}
    \label{fig:ins}
    \vspace{-1.5em}
\end{wrapfigure}

Core institutions of China and the United States mainly maintained a dominant position across the period, as shown in \cref{fig:ins_count}. Tsinghua University has long been at the top in terms of the number of publications, demonstrating the stability of its research output and its ability to continuously accumulate. However, there were annual competitive dynamic changes in the rankings between industrial giants and universities. In terms of the cooperation network, as shown in \cref{fig:ins}, domestic universities have active collaborations with industrial giants such as Alibaba, Tencent, and Huawei, showing obvious regional characteristics. At the same time, Chinese and American institutions still dominate the cooperation network, forming a "core-multipoint" cooperation pattern. This shift in research paradigms, driven by LLMs and Benchmark, is also profoundly reflected in the research strategies of different institutions. In the AI field, the academic community and the industrial sector have their own focuses and develop in a complementary manner. This is not simply a difference in interests, but rather reflects the strategic differentiation in the ecological niche of scientific research:
\begin{itemize}
    \item \textbf{Academic institutions typically focus on basic research and cutting-edge algorithm exploration}, with their strategic priorities lying in efficiency improvement, robustness assurance, and continuous investigation of theoretical mechanisms, aiming to advance the democratization of AI technology. For instance, Tsinghua University emphasizes directions such as knowledge distillation, graph neural networks, adversarial training, domain adaptation, and model generalization, while Carnegie Mellon University demonstrates strong performance in areas like robotic grasping and manipulation strategies and causal discovery.
    \item \textbf{Industrial institutions lean more toward application-oriented research and theoretical studies related to technology deployment.} Leveraging their advantages in data and computing resources, they strive to address issues such as system efficiency and real-time performance in large-scale model deployment (Scaling-up) and commercial applications, establishing a complete AI ecosystem and technological moat. For example, Microsoft maintains leadership in LLMs inference, code intelligence, mathematical reasoning, and retrieval-augmented generation, while Google focuses more on application and system-level topics, including speech recognition, machine translation, multi-task learning, and federated learning.
\end{itemize}

\section{Retrieval and Analysis}

To evaluate the effectiveness of our retrieval-augmented LLM framework for semantic understanding and literature summarization, we selected the Top-20 topics based on paper volume and diversity across 22 AI conferences from 2020 to 2025. For each topic, papers were processed using two methods: (i) direct generation with ChatGPT-5, and (ii) retrieval-augmented generation leveraging our $ResearchDB$. ChatGPT-5 is used as a representative baseline, while the retrieval-augmented approach can be applied to other LLMs and assessed in future work. Evaluation was performed by five doctoral students along five dimensions:

\begin{itemize}
    \item \textbf{Accuracy}: Consistency of the generated content with the original papers;
    \item \textbf{Coverage}: Inclusion of core content and major sub-directions within the topic;
    \item \textbf{Novelty}: Ability to capture recently emerging directions or methods;
    \item \textbf{Readability}: Clarity and logical coherence of the text;
    \item \textbf{Usefulness}: Facilitation of rapid understanding and insight into potential research trends.
\end{itemize}
\begin{wrapfigure}{r}{0.45\textwidth}
    \centering
      \vspace{-1em}
    \includegraphics[width=0.8\linewidth]{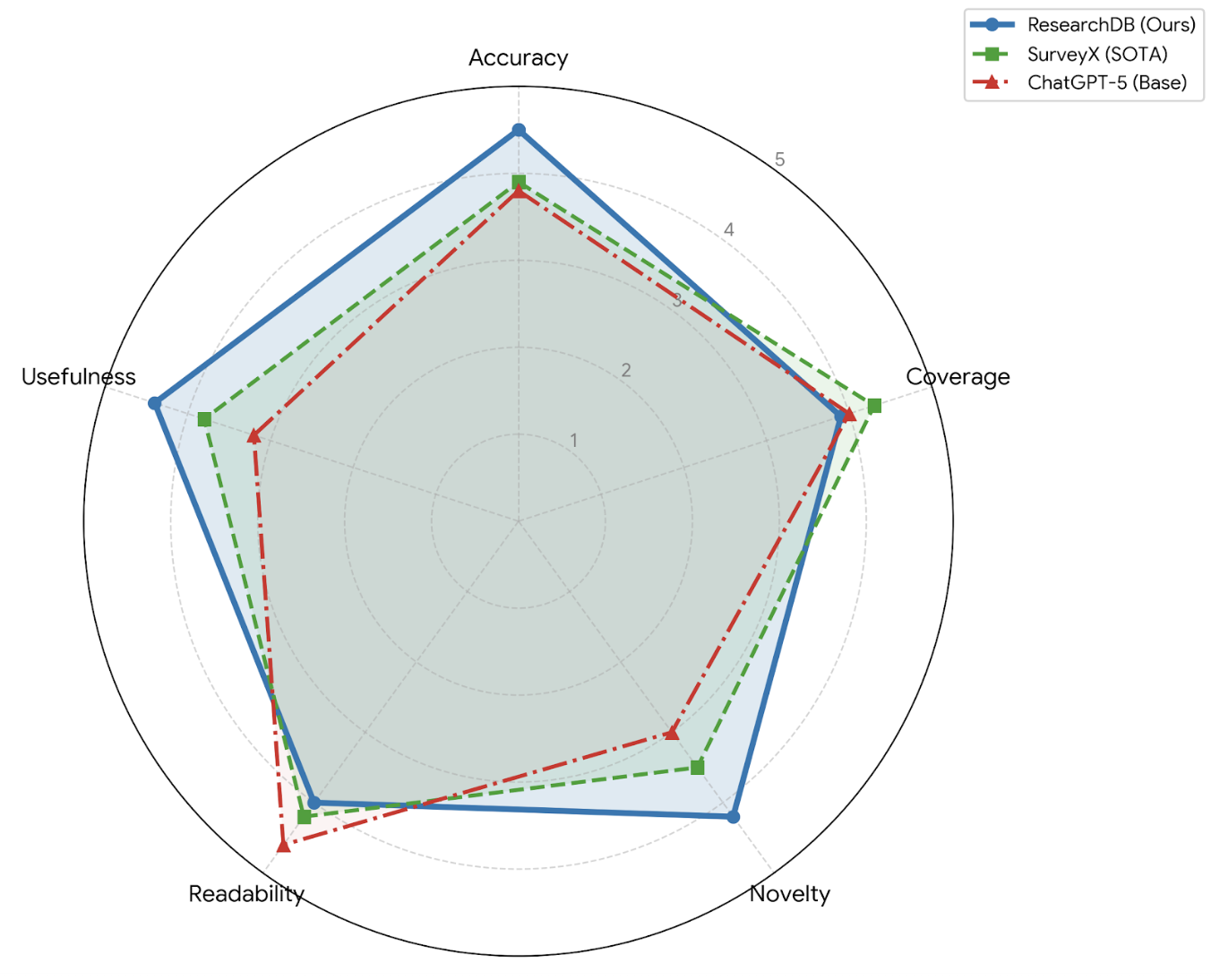}
     \vspace{-0.5em}
    \caption{
    Performance comparison of $ResearchDB$ against the SurveyX method and ChatGPT-5 across five evaluation dimensions, based on expert scoring.
    }
    \label{fig:lidar}
    \vspace{-1em}
\end{wrapfigure} 
Each criterion was scored on a 5-point Likert scale independently by all annotators. Inter-annotator agreement was measured using Cohen’s kappa to ensure consistency.

The evaluation results in \cref{fig:lidar} reveal that our proposed framework outperforms the baseline methods across several critical dimensions. Compared to direct generation by ChatGPT-5, our approach achieves significantly higher Accuracy and Usefulness. This improvement is primarily attributed to our intent-driven hierarchical retrieval, which grounds the model's responses in structured evidence extracted from specific content fields. 
While SurveyX\cite{liang2025surveyx} focuses on general automation, our framework excels in capturing technical specifics through weighted multi-field search. This evidence-grounded approach prioritizes high-fidelity sections (e.g., methods, datasets), ensuring superior reliability and traceability over unconstrained LLM generation , which is essential for high-stakes research analysis

\subsection{Ablation Study}
\begin{table}[t]
\centering
\small  
\setlength{\tabcolsep}{5pt}  
\caption{Ablation results on retrieval performance based on expert scores.}
\label{tab:ablation}
\begin{tabularx}{\linewidth}{l *{5}{>{\centering\arraybackslash}X}}
\toprule
\textbf{Variant} & \textbf{Acc.} & \textbf{Cov.} & \textbf{Nov.} & \textbf{Read.} & \textbf{Use.} \\
\midrule
w/o MF           & 4.12          & 3.85          & 4.00          & 4.10           & 3.90          \\
w/o DW           & 3.86          & 3.82          & 3.70          & 4.05           & 3.75          \\
SF               & 3.70          & 3.10          & 3.20          & 4.20           & 3.25          \\
\textbf{ResearchDB} & \textbf{4.50} & \textbf{3.90} & \textbf{4.20} & \textbf{4.00} & \textbf{4.40} \\
\bottomrule
\end{tabularx}
\end{table}
To evaluate the contribution of each core component in our hierarchical retrieval pipeline, we conduct an ablation study by comparing the full ResearchDB framework against three degraded variants consisting of a configuration without metadata filtering (w/o MF), a version without dynamic weighting (w/o DW) using a uniform weight distribution, and a single-field baseline (SF) that restricts semantic search to abstracts only. As demonstrated by the experimental results in \cref{tab:ablation}, the full framework consistently achieves the highest performance across all metrics, with the most significant drop in accuracy occurring when the intent-driven dynamic weighting mechanism is disabled. Specifically, the removal of dynamic weighting prevents the system from effectively prioritizing task-relevant fields; Furthermore, the exclusion of metadata filtering leads to a noticeable decrease in precision as the system fails to exclude papers from irrelevant venues or timeframes that share similar semantic embeddings. Finally, the inferior performance of the single-field baseline confirms that relying solely on abstracts is insufficient for deep knowledge profiling, as essential information regarding datasets and limitations is often omitted or overly generalized in paper summaries, proving that the integration of structured multidimensional fields with adaptive weights is vital for high-fidelity literature analysis.

\section{Conclusion} \label{sec:conclusion}

We present a multidimensional knowledge profiling framework that integrates topic clustering, LLM-based semantic parsing, and hierarchical retrieval to analyze over 100K AI publications. By capturing topic dynamics, dataset trends, and institutional patterns, the framework enables evidence-driven discovery of methodological shifts and technical paradigms. Our approach provides both a macro-level landscape overview and fine-grained, topic-level insights, effectively supporting systematic trend analysis and research planning.

\noindent\textbf{Limitations and Impact.} While our study is limited to major English-language AI conferences and subject to potential LLM parsing biases, it offers a transparent, evidence-based perspective on research evolution. Future work will aim to improve domain-specific parsing accuracy and extend the framework to a broader range of scientific venues.

\bibliography{april_aigc}

@String(AAAI = {AAAI})

@article{zupic2015bibliometric,
  title={Bibliometric methods in management and organization},
  author={Zupic, Ivan and {\v{C}}ater, Toma{\v{z}}},
  journal={Organizational research methods},
  volume={18},
  number={3},
  pages={429--472},
  year={2015},
  publisher={Sage Publications Sage CA: Los Angeles, CA}
}

@article{donthu2021conduct,
  title={How to conduct a bibliometric analysis: An overview and guidelines},
  author={Donthu, Naveen and Kumar, Satish and Mukherjee, Debmalya and Pandey, Nitesh and Lim, Weng Marc},
  journal={Journal of business research},
  volume={133},
  pages={285--296},
  year={2021},
  publisher={Elsevier}
}

@article{chen2006citespace,
  title={CiteSpace II: Detecting and visualizing emerging trends and transient patterns in scientific literature},
  author={Chen, Chaomei},
  journal={Journal of the American Society for information Science and Technology},
  volume={57},
  number={3},
  pages={359--377},
  year={2006},
  publisher={Wiley Online Library}
}

@article{van2010software,
  title={Software survey: VOSviewer, a computer program for bibliometric mapping},
  author={Van Eck, Nees and Waltman, Ludo},
  journal={scientometrics},
  volume={84},
  number={2},
  pages={523--538},
  year={2010},
  publisher={Akad{\'e}miai Kiad{\'o}, co-published with Springer Science+ Business Media BV~…}
}

@article{aria2017bibliometrix,
  title={bibliometrix: An R-tool for comprehensive science mapping analysis},
  author={Aria, Massimo and Cuccurullo, Corrado},
  journal={Journal of informetrics},
  volume={11},
  number={4},
  pages={959--975},
  year={2017},
  publisher={Elsevier}
}

@article{blei2012probabilistic,
  title={Probabilistic topic models},
  author={Blei, David M},
  journal={Communications of the ACM},
  volume={55},
  number={4},
  pages={77--84},
  year={2012},
  publisher={ACM New York, NY, USA}
}

@article{blei2003latent,
  title={Latent dirichlet allocation},
  author={Blei, David M and Ng, Andrew Y and Jordan, Michael I},
  journal={Journal of machine learning research},
  volume={3},
  number={Jan},
  pages={993--1022},
  year={2003}
}

@inproceedings{alsumait2009topic,
  title={Topic significance ranking of LDA generative models},
  author={AlSumait, Loulwah and Barbar{\'a}, Daniel and Gentle, James and Domeniconi, Carlotta},
  booktitle={Joint European conference on machine learning and knowledge discovery in databases},
  pages={67--82},
  year={2009},
  organization={Springer}
}

@article{chang2009reading,
  title={Reading tea leaves: How humans interpret topic models},
  author={Chang, Jonathan and Gerrish, Sean and Wang, Chong and Boyd-Graber, Jordan and Blei, David},
  journal={Advances in neural information processing systems},
  volume={22},
  year={2009}
}

@inproceedings{chuang2013topic,
  title={Topic model diagnostics: Assessing domain relevance via topical alignment},
  author={Chuang, Jason and Gupta, Sonal and Manning, Christopher and Heer, Jeffrey},
  booktitle={International conference on machine learning},
  pages={612--620},
  year={2013},
  organization={PMLR}
}

@article{grootendorst2022bertopic,
  title={BERTopic: Neural topic modeling with a class-based TF-IDF procedure},
  author={Grootendorst, Maarten},
  journal={arXiv preprint arXiv:2203.05794},
  year={2022}
}

@article{reimers2019sentence,
  title={Sentence-bert: Sentence embeddings using siamese bert-networks},
  author={Reimers, Nils and Gurevych, Iryna},
  journal={arXiv preprint arXiv:1908.10084},
  year={2019}
}

@inproceedings{devlin2019bert,
  title={Bert: Pre-training of deep bidirectional transformers for language understanding},
  author={Devlin, Jacob and Chang, Ming-Wei and Lee, Kenton and Toutanova, Kristina},
  booktitle={Proceedings of the 2019 conference of the North American chapter of the association for computational linguistics: human language technologies, volume 1 (long and short papers)},
  pages={4171--4186},
  year={2019}
}

@article{ccelikten2025topic,
  title={Topic modeling through rank-based aggregation and LLMs: An approach for AI and human-generated scientific texts},
  author={{\c{C}}elikten, Tu{\u{g}}ba and Onan, Aytu{\u{g}}},
  journal={Knowledge-Based Systems},
  volume={314},
  pages={113219},
  year={2025},
  publisher={Elsevier}
}

@article{kostikova2025lllms,
  title={LLLMs: A Data-Driven Survey of Evolving Research on Limitations of Large Language Models},
  author={Kostikova, Aida and Wang, Zhipin and Bajri, Deidamea and P{\"u}tz, Ole and Paa{\ss}en, Benjamin and Eger, Steffen},
  journal={arXiv preprint arXiv:2505.19240},
  year={2025}
}

@article{diaz2025k,
  title={k-LLMmeans: Summaries as Centroids for Interpretable and Scalable LLM-Based Text Clustering},
  author={Diaz-Rodriguez, Jairo},
  journal={arXiv e-prints},
  pages={arXiv--2502},
  year={2025}
}

@inproceedings{lam2024concept,
  title={Concept induction: Analyzing unstructured text with high-level concepts using lloom},
  author={Lam, Michelle S and Teoh, Janice and Landay, James A and Heer, Jeffrey and Bernstein, Michael S},
  booktitle={Proceedings of the 2024 CHI Conference on Human Factors in Computing Systems},
  pages={1--28},
  year={2024}
}

@article{si2024can,
  title={Can llms generate novel research ideas? a large-scale human study with 100+ nlp researchers},
  author={Si, Chenglei and Yang, Diyi and Hashimoto, Tatsunori},
  journal={arXiv preprint arXiv:2409.04109},
  year={2024}
}

@article{scherbakov2025emergence,
  title={The emergence of large language models as tools in literature reviews: a large language model-assisted systematic review},
  author={Scherbakov, Dmitry and Hubig, Nina and Jansari, Vinita and Bakumenko, Alexander and Lenert, Leslie A},
  journal={Journal of the American Medical Informatics Association},
  volume={32},
  number={6},
  pages={1071--1086},
  year={2025},
  publisher={Oxford University Press}
}

@article{brown2020language,
  title={Language models are few-shot learners},
  author={Brown, Tom and Mann, Benjamin and Ryder, Nick and Subbiah, Melanie and Kaplan, Jared D and Dhariwal, Prafulla and Neelakantan, Arvind and Shyam, Pranav and Sastry, Girish and Askell, Amanda and others},
  journal={Advances in neural information processing systems},
  volume={33},
  pages={1877--1901},
  year={2020}
}

@article{achiam2023gpt,
  title={Gpt-4 technical report},
  author={Achiam, Josh and Adler, Steven and Agarwal, Sandhini and Ahmad, Lama and Akkaya, Ilge and Aleman, Florencia Leoni and Almeida, Diogo and Altenschmidt, Janko and Altman, Sam and Anadkat, Shyamal and others},
  journal={arXiv preprint arXiv:2303.08774},
  year={2023}
}

@article{comanici2025gemini,
  title={Gemini 2.5: Pushing the frontier with advanced reasoning, multimodality, long context, and next generation agentic capabilities},
  author={Comanici, Gheorghe and Bieber, Eric and Schaekermann, Mike and Pasupat, Ice and Sachdeva, Noveen and Dhillon, Inderjit and Blistein, Marcel and Ram, Ori and Zhang, Dan and Rosen, Evan and others},
  journal={arXiv preprint arXiv:2507.06261},
  year={2025}
}

@article{yang2025qwen3,
  title={Qwen3 technical report},
  author={Yang, An and Li, Anfeng and Yang, Baosong and Zhang, Beichen and Hui, Binyuan and Zheng, Bo and Yu, Bowen and Gao, Chang and Huang, Chengen and Lv, Chenxu and others},
  journal={arXiv preprint arXiv:2505.09388},
  year={2025}
}

@article{touvron2023llama,
  title={Llama: Open and efficient foundation language models},
  author={Touvron, Hugo and Lavril, Thibaut and Izacard, Gautier and Martinet, Xavier and Lachaux, Marie-Anne and Lacroix, Timoth{\'e}e and Rozi{\`e}re, Baptiste and Goyal, Naman and Hambro, Eric and Azhar, Faisal and others},
  journal={arXiv preprint arXiv:2302.13971},
  year={2023}
}

@article{lala2023paperqa,
  title={Paperqa: Retrieval-augmented generative agent for scientific research},
  author={L{\'a}la, Jakub and O'Donoghue, Odhran and Shtedritski, Aleksandar and Cox, Sam and Rodriques, Samuel G and White, Andrew D},
  journal={arXiv preprint arXiv:2312.07559},
  year={2023}
}

@article{skarlinski2024language,
  title={Language agents achieve superhuman synthesis of scientific knowledge},
  author={Skarlinski, Michael D and Cox, Sam and Laurent, Jon M and Braza, James D and Hinks, Michaela and Hammerling, Michael J and Ponnapati, Manvitha and Rodriques, Samuel G and White, Andrew D},
  journal={arXiv preprint arXiv:2409.13740},
  year={2024}
}

@article{besrour2025squai,
  title={SQuAI: Scientific Question-Answering with Multi-Agent Retrieval-Augmented Generation},
  author={Besrour, Ines and He, Jingbo and Schreieder, Tobias and F{\"a}rber, Michael},
  journal={arXiv preprint arXiv:2510.15682},
  year={2025}
}

@article{gao2023retrieval,
  title={Retrieval-augmented generation for large language models: A survey},
  author={Gao, Yunfan and Xiong, Yun and Gao, Xinyu and Jia, Kangxiang and Pan, Jinliu and Bi, Yuxi and Dai, Yixin and Sun, Jiawei and Wang, Haofen and Wang, Haofen},
  journal={arXiv preprint arXiv:2312.10997},
  volume={2},
  number={1},
  year={2023}
}

@article{cheng2025survey,
  title={A survey on knowledge-oriented retrieval-augmented generation},
  author={Cheng, Mingyue and Luo, Yucong and Ouyang, Jie and Liu, Qi and Liu, Huijie and Li, Li and Yu, Shuo and Zhang, Bohou and Cao, Jiawei and Ma, Jie and others},
  journal={arXiv preprint arXiv:2503.10677},
  year={2025}
}

@article{nguyen2025ma,
  title={MA-RAG: Multi-Agent Retrieval-Augmented Generation via Collaborative Chain-of-Thought Reasoning},
  author={Nguyen, Thang and Chin, Peter and Tai, Yu-Wing},
  journal={arXiv preprint arXiv:2505.20096},
  year={2025}
}

@article{wu2025structure,
  title={Structure-R1: Dynamically Leveraging Structural Knowledge in LLM Reasoning through Reinforcement Learning},
  author={Wu, Junlin and Zhong, Xianrui and Sun, Jiashuo and Li, Bolian and Jin, Bowen and Han, Jiawei and Zeng, Qingkai},
  journal={arXiv preprint arXiv:2510.15191},
  year={2025}
}

@article{singh2025agentic,
  title={Agentic retrieval-augmented generation: A survey on agentic rag},
  author={Singh, Aditi and Ehtesham, Abul and Kumar, Saket and Khoei, Tala Talaei},
  journal={arXiv preprint arXiv:2501.09136},
  year={2025}
}

@article{li2025towards,
  title={Towards agentic rag with deep reasoning: A survey of rag-reasoning systems in llms},
  author={Li, Yangning and Zhang, Weizhi and Yang, Yuyao and Huang, Wei-Chieh and Wu, Yaozu and Luo, Junyu and Bei, Yuanchen and Zou, Henry Peng and Luo, Xiao and Zhao, Yusheng and others},
  journal={arXiv preprint arXiv:2507.09477},
  year={2025}
}

@article{wang2024autosurvey,
  title={Autosurvey: Large language models can automatically write surveys},
  author={Wang, Yidong and Guo, Qi and Yao, Wenjin and Zhang, Hongbo and Zhang, Xin and Wu, Zhen and Zhang, Meishan and Dai, Xinyu and Wen, Qingsong and Ye, Wei and others},
  journal={Advances in neural information processing systems},
  volume={37},
  pages={115119--115145},
  year={2024}
}

@article{liang2025surveyx,
  title={Surveyx: Academic survey automation via large language models},
  author={Liang, Xun and Yang, Jiawei and Wang, Yezhaohui and Tang, Chen and Zheng, Zifan and Song, Shichao and Lin, Zehao and Yang, Yebin and Niu, Simin and Wang, Hanyu and others},
  journal={arXiv preprint arXiv:2502.14776},
  year={2025}
}

@inproceedings{yan2025surveyforge,
  title={Surveyforge: On the outline heuristics, memory-driven generation, and multi-dimensional evaluation for automated survey writing},
  author={Yan, Xiangchao and Feng, Shiyang and Yuan, Jiakang and Xia, Renqiu and Wang, Bin and Bai, Lei and Zhang, Bo},
  booktitle={Proceedings of the 63rd Annual Meeting of the Association for Computational Linguistics (Volume 1: Long Papers)},
  pages={12444--12465},
  year={2025}
}

@article{nguye2025surveyg,
  title={SurveyG: A Multi-Agent LLM Framework with Hierarchical Citation Graph for Automated Survey Generation},
  author={Nguye, Minh-Anh and Nguyen, Minh-Duc and Lan, Nguyen Thi Ha and Dang, Kieu Hai and Dong, Nguyen Tien and Dung, Le Duy},
  journal={arXiv preprint arXiv:2510.07733},
  year={2025}
}

@article{wang2024mineru,
  title={Mineru: An open-source solution for precise document content extraction},
  author={Wang, Bin and Xu, Chao and Zhao, Xiaomeng and Ouyang, Linke and Wu, Fan and Zhao, Zhiyuan and Xu, Rui and Liu, Kaiwen and Qu, Yuan and Shang, Fukai and others},
  journal={arXiv preprint arXiv:2409.18839},
  year={2024}
}

@article{guo2025deepseek,
  title={Deepseek-r1: Incentivizing reasoning capability in llms via reinforcement learning},
  author={Guo, Daya and Yang, Dejian and Zhang, Haowei and Song, Junxiao and Zhang, Ruoyu and Xu, Runxin and Zhu, Qihao and Ma, Shirong and Wang, Peiyi and Bi, Xiao and others},
  journal={arXiv preprint arXiv:2501.12948},
  year={2025}
}

@article{mcinnes2018umap,
  title={Umap: Uniform manifold approximation and projection for dimension reduction},
  author={McInnes, Leland and Healy, John and Melville, James},
  journal={arXiv preprint arXiv:1802.03426},
  year={2018}
}

@article{mcinnes2017hdbscan,
  title={hdbscan: Hierarchical density based clustering.},
  author={McInnes, Leland and Healy, John and Astels, Steve and others},
  journal={J. Open Source Softw.},
  volume={2},
  number={11},
  pages={205},
  year={2017}
}

@inproceedings{adavideorag,
  title={AdaVideoRAG: Omni-Contextual Adaptive Retrieval-Augmented Efficient Long Video Understanding.},
  author={Xue, Zhucun and Zhang, Jiangning and Xie, Xurong and Cai, Yuxuan and Liu, Yong and Li, Xiangtai and Tao, Dacheng},
  booktitle={NeurIPS},
  year={2025}
}

@inproceedings{adakd,
  title={LLM-Oriented Token-Adaptive Knowledge Distillation},
  author={Xie, Xurong and Xue, Zhucun and Wu, Jiafu and Li, Jian and Wang, Yabiao and Hu, Xiaobin and Liu, Yong and Zhang, Jiangning},
  booktitle={AAAI},
  year={2026}
}

\clearpage
\renewcommand\thefigure{A\arabic{figure}}
\renewcommand\thetable{A\arabic{table}}  
\renewcommand\theequation{A\arabic{equation}}
\setcounter{equation}{0}
\setcounter{table}{0}
\setcounter{figure}{0}
\appendix
\section*{Appendix}
\textbf{Overview}
\label{appendix}

The appendix presents the following sections to strengthen the main manuscript:

\begin{itemize}
\item[—] We provide more details in the Supplementary Materials for a better understanding of our work. 
\item[—] \textbf{Sec.}~A. $ResearchDB$ Source .  
\item[—] \textbf{Sec.}~B. Implementation and Reproducibility. 
\item[—] \textbf{Sec.}~C. Topic Clustering Results. 
\item[—] \textbf{Sec.}~D. Retrieval System Evaluation. 
\end{itemize}

\section{$ResearchDB$ Source } \label{sec:dataset}
This appendix details the source, scale, and temporal distribution of the underlying database, as well as the multidimensional data schema used to construct the $ResearchDB$ knowledge graph.

\noindent\textbf{Database Source and Distribution.}
The $ResearchDB$ comprises over 100,000 papers from 22 top-tier conferences and journals published between 2020 and 2025. This broad selection ensures comprehensive coverage across key AI and Computer Science sub-domains, as shown in \cref{Tab: conv}.

\begin{table*}[!htp]
\centering
\caption{The 22 major venues in the ResearchDB corpus, categorized by domain, including their abbreviations and total paper count (2020-2025).}
\label{tab:corpus_list_compact}
\resizebox{\textwidth}{!}{%
\begin{tabular}{llcc}
\toprule
\textbf{Domain} & \textbf{Venue Name} & \textbf{Abbr.} & \textbf{Total Papers} \\
\midrule
\multirow{3}{*}{\shortstack{Computer\\ Vision}} & Conference on Computer Vision and Pattern Recognition & CVPR & 11436 \\
& International Conference on Computer Vision & ICCV & 3768 \\
& European Conference on Computer Vision & ECCV & 5389 \\
\midrule
\multirow{5}{*}{\shortstack{NLP \&\\ Speech}} & Annual Meeting of the Association for Computational Linguistics & ACL & 10595 \\
& Conference on Empirical Methods in NLP & EMNLP & 8703 \\
& North American Chapter of the ACL & NAACL & 3625 \\
& International Conf. on Comp. Linguistics (Modeling) & COLM & 299 \\
& International Speech Communication Association & INTERSPEECH & 5351 \\
\midrule
\multirow{9}{*}{\shortstack{Core ML\\ \& AI}} & Conference on Neural Information Processing Systems & NeurIPS & 12726 \\
& International Conference on Machine Learning & ICML & 10199 \\
& International Conference on Learning Representations & ICLR & 9512 \\
& Association for the Advancement of Artificial Intelligence & AAAI & 13575 \\
& International Joint Conference on Artificial Intelligence & IJCAI & 4263 \\
& Conference on Uncertainty in Artificial Intelligence & UAI & 1245 \\
& Conference on Robot Learning & CoRL & 814 \\
& International Workshop on Spoken Language Translation & IWSLT & 187 \\
& Conference on Learning Theory & COLT & 755 \\
\midrule
\multirow{5}{*}{\shortstack{Systems\\ \& Security}} & Machine Learning and Systems & MLSYS & 220 \\
& USENIX Symp. on Networked Systems Design and Implementation & OSDI & 258 \\
& Network and Distributed System Security Symposium & NDSS & 409 \\
& USENIX Conf. on File and Storage Technologies & USENIX-Fast & 165\\
& USENIX Security Symposium & USENIX-Sec & 988 \\
\midrule
\multicolumn{3}{r}{\textbf{Overall Total}} & \textbf{$\approx 104482$} \\ 
\bottomrule
\end{tabular}}
\label{Tab: conv}
\end{table*}

\section{Implementation and Reproducibility.}\label{sec:llm}
This section details the specific parameters and configurations for the two core components of our large-scale multidimensional knowledge profiling framework: LLM-assisted knowledge parsing and multidimensional topic clustering, ensuring full reproducibility of our methodology.

\noindent\textbf{LLM Knowledge Parsing Prompts.}
We utilized the Deepseek-R1-32B model for semantic parsing of full paper texts to populate the $ResearchDB$ schema (as documented in \cref{sec:dataset}). The parsing fidelity hinges on a carefully designed, JSON constrained prompt template.
Below is the primary prompt template used for extracting core technical information (e.g., methods, architecture, training\_setup).

\begin{lstlisting}

You are a scientific paper analysis assistant.

Your task is to read the input paper content (in Markdown format), and output a structured summary in JSON format according to the schema below. 

Output JSON schema:
{
  "abstract_summary": "",
  "keywords": [],
  "keywords_explanation": {},
  "methods": "",
  "architecture": "",
  "loss_function": "",
  "training_setup": "",
  "gpu_info": ""
  "datasets": [],
  "metrics": [],
  "datasets_metrics_mapping": {},
  "problem_statement": "",
  "contributions": [],
  "novelty_type": "",
  "experiments": "",
  "results_summary": "",
  "limitations": [],
  "future_work": [],
  "trend_insight": "",
  "field_positioning": ""
  "institution": []
}

Instructions:
1. **Keywords**: Extract 3-10 keywords directly from the paper text. Generate a short explanation for each keyword using LLM summarization.
2. **abstract_summary**: Provide a compressed version of the paper's abstract. Summarize, do not copy verbatim.
3. **Methods / Architecture / Loss / Training**: Summarize core method(s), model architecture, loss function, and training setup.
4. **Datasets / Metrics / Mapping**: List datasets, corresponding metrics, and mapping between datasets and metrics. Use empty lists/dictionaries if missing.
5. **Problem Statement**: Summarize the research problem in 1-2 sentences.
6. **Contributions**: Summarize the core contributions.
7. **Novelty Type**: Infer innovation type:
   - Algorithm / Model
   - Theory / Analysis
   - Benchmark / Dataset
   - Application / System
   - Methodological Improvement
8. **Experiments / Results Summary**: Summarize experiments and main results. Include ablation studies if present.
9. **Limitations / Future Work**: Summarize limitations and future research directions.
10. **Trend Insight**: Provide trend level insights based on the paper.
11. **Field Positioning**: Infer the paper's position in its research field:
    - Foundational Work
    - Methodological Innovation
    - Benchmark / Dataset Contribution
    - Application Validation
    - Trend Extension
12. **Institution**:Extract all author affiliations from the provided.
13. **Gpu_info**:Obtain the gpu resources used in the article, <total\_gpu>*<GPU_MODEL>*<training_time>.
\end{lstlisting}

\noindent\textbf{LLM Intent Recognition Prompts.}
The core of our retrieval system's ability to effectively handle complex, multi-dimensional queries lies in an LLM-based intent recognition module. The goal of this module is to convert users' natural language queries into a format that strictly matches the data schema of the $ResearchDB$ knowledge database.
\begin{lstlisting}
You are a comprehensive query analysis and search planning assistant for a scientific literature knowledge base.

Your primary task is to parse the user's natural language query into a single, structured JSON object that captures all metadata constraints, technical requirements, and generates an optimized plan for vector search.

Output Schema:
{
  "conference": [],
  "year": [],
  "paper_name": [],
  "authors": [],
  "keywords": [],
  "keywords_explanation": {},
  "abstract_summary": "",
  "methods": "",
  "architecture": "",
  "loss_function": "",
  "datasets": [],
  "metrics": [],
  "vector_search_plan": []
}

Parsing Rules:
1. **Conference**: Must the range of the following fixed options: [AAAI, ACL, COLM, COLT, CoRL, CVPR, ECCV, EMNLP, ICCV, ICLR, ICML, IJCAI, INTERSPEECH, IWSLT, MLSYS, NAACL, NDSS, NeurIPS, OSDI, UAI, USENIX-Fast, USENIX-Sec]. If the query does not mention a conference, leave as []. 
2. **Year**: Extract the range of year contained in the query.
3. **Authors**: If the query explicitly mentions author names, extract them into a list. Otherwise, leave empty. 
4. **Paper Name**: Extract all explicitly mentioned paper titles. Output as a list. Use partial matching if necessary.
5. **Keywords**: Extract only technical terms explicitly mentioned in query. Use `keywords_explanation` to briefly explain each.
6. **Abstract Summary**: One concise sentence describing technical focus from query.
6. **methods**:  
   - Output the **complete method or approach description**, including its application domain if present.  
   - Must describe both the **technique** and its **application scope**, not just a keyword.  
   - If multiple distinct methods are present, include all as a list.  
   - Do NOT shorten into keywords.  
7. ** Architecture / Loss Function / Datasets / Metrics**: Extract explicitly if mentioned; do not hallucinate.
8. **Vector Search Plan**:  
   - Decide which fields to include in the search (`abstract_summary`, `methods`, `keywords`, `datasets`, `metrics`, `architecture`, `loss_function`, etc.).  
   - Only include fields that have non-empty content in the JSON output.  
   - Assign a `weight` for each field between 0 and 1 based on **query emphasis**.  
   - The weights **must be normalized such that their sum equals 1.0**.  
   - Fields are **not mutually exclusive**; e.g., methods can also appear in keywords.  
   - Weights should reflect relative importance of each field **according to the query intent**, not by default priority.
\end{lstlisting}

\noindent\textbf{ResearchDB Topic Clustering Configuration.}
The multidimensional topic structure of $ResearchDB$ was generated through a robust embedding-based pipeline. We first transformed the paper abstracts and key textual segments into semantic vectors using the all-MiniLM-L6-v2 model. These high dimensional embeddings were then processed by UMAP for noise reduction and projection into a lower dimensional space, specifically configured with $n\_components=40$, $n\_neighbors=120$, and $min\_dist=0.08$ (with a fixed $random\_state=42$ for reproducibility). UMAP utilized the cosine metric, which aligns with the semantic nature of the embeddings. Clustering was performed using HDBSCAN, which is optimized for identifying clusters of varying densities; the algorithm was set with a significantly smaller $min\_cluster\_size=50$ and $min\_samples=1$, employing the \texttt{eom} (Excess of Mass) cluster selection method. Topic representation was established using C-TF-IDF, configured to filter vocabulary using $\text{MAX\_DF\_PERCENT}=0.9$ and a $\text{MIN\_DF\_ABS\_FLOOR}=50$, utilizing an n-gram range of (1, 2). The full hierarchical structure, derived from the HDBSCAN tree and refined by an external LLM (ChatGPT-5) for descriptive naming, resulted in the fine-grained multi-level taxonomy of over 300 topics presented in \cref{sec: topic}.

\begin{table*}[htbp]
\centering
\small
\setlength{\tabcolsep}{4pt}
\caption{Sample list of fine-grained topics extracted from $ResearchDB$, illustrating the specificity and coverage of the taxonomy.}
\label{tab:tiny}
\begin{tabularx}{\textwidth}{l c >{\raggedright\arraybackslash}p{3.8cm} X}
\toprule
\textbf{Domain} & \textbf{ID} & \textbf{Topic Name} & \textbf{Representative Keywords} \\
\midrule
Optimization Theory & 140 & Approximation and Fair Clustering Algorithms & correlation clustering, approximation ratio, $\ell_p$ norm \\
Computer Vision & 088 & Efficient ViT Training and Pruning & vision transformer, token pruning, dynamic exit\\
Natural Language Processing & 215 & Controllable Text Generation via Latent Space & VAE, latent variable, attribute control, language model \\
Core ML & 042 & Causal Inference for Robustness and Generalization & do-calculus, invariant prediction, domain generalization \\
Systems \& Security & 291 & Hardware Acceleration for Neural Network Inference & FPGA, ASIC, low-bit quantization, hardware optimization \\
\bottomrule
\end{tabularx}
\label{tab: topic_list}
\end{table*}

\section{Topic Clustering Results.}\label{sec: topic}
This section presents the complete results of the fine-grained topic taxonomy for ResearchDB. By combining sentence-transformers embeddings, UMAP dimensionality reduction, and the HDBSCAN clustering algorithm, we successfully identified and named 324 independent, high-precision research topics from the corpus of over 100,000 papers. The HDBSCAN configuration ensures that each topic represents a sufficiently large yet highly homogenous set of papers, guaranteeing the fine granularity and specificity of the topics.

\begin{figure}
    \centering
    \includegraphics[width=1\linewidth]{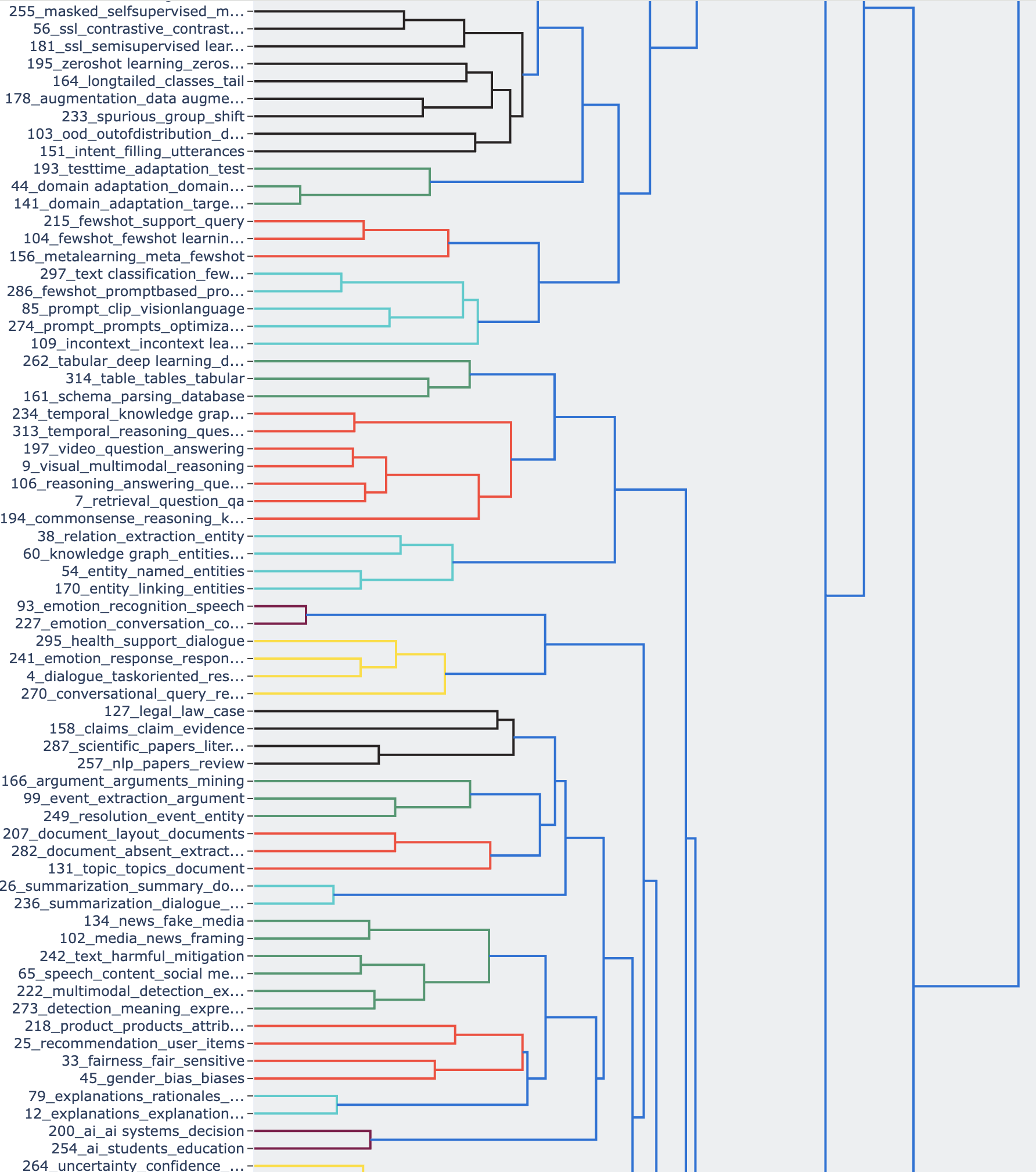}
    \caption{Hierarchical topic tree structure. }
    \label{fig:Hierarchical}
    
\end{figure}

\noindent\textbf{Hierarchical Topic Tree Structure.}
A key advantage of the HDBSCAN algorithm is its inherent ability to produce a Hierarchical Structure. We leverage this structure as an organizational framework, providing a non-strict view of clustering relationships above the 324 fine-grained topics. This tree structure is the basis for our hierarchical retrieval system, allowing users to navigate from macro-level concepts (near the root) down to specific fine-grained topics (at the leaves), as shown in \cref{fig:Hierarchical}.

\noindent\textbf{Topic Naming and Examples.}
The naming of each fine-grained topic underwent strict quality control: C-TF-IDF was used to extract representative keywords, which were then analyzed and refined by ChatGPT-5 based on paper abstracts within the cluster. This process generated highly semantically clear Topic Names (e.g., 140\_Approximation and Fair Clustering Algorithms). Every paper in $ResearchDB$ is uniquely tagged with such a topic name.  \cref{tab: topic_list} below presents a sample of the most representative fine-grained topics. 
We also have selected topics with ids ranging from 1 to 20 and listed them as \cref{tab: topic_detail}

\begin{table*}[htp]
\centering
\scriptsize 

\caption{Twenty Sample Fine-Grained Topics from $ResearchDB$, Illustrating the Taxonomy's Scope and Detail.}

\begin{tabularx}{\textwidth}{@{}rp{2.6cm}X@{}}
\toprule
\textbf{ID} & \textbf{Topic Name} & \textbf{Summary} \\
\midrule
0 & Graph Neural Networks and Representation Learning & Research advancing spectral and transformer-based GNNs, self-supervised and pretraining strategies, and scalable training to learn robust node and graph representations under heterophily, temporal dependencies, and distribution shifts. \\
\midrule
1 & Efficient Long-Context Attention and Memory & Research on extending and efficiently leveraging long contexts in transformers via improved positional encodings, attention approximations, adaptive computation, and memory augmentation, enabling length generalization and scalable training and inference. \\
\midrule
2 & Adversarial Attacks and Robustness & Research advances adversarial example generation and transferability across modalities and tasks while developing training, architectural, certification, and benchmarking methods to improve and assess deep models’ robustness against digital, physical, and black-box attacks. \\
\midrule
3 & AI-Driven Molecular and Protein Design & Research leveraging structure- and sequence-informed machine learning, including graph and equivariant models, to predict properties, model interactions, and generate designs for molecules, proteins, and materials. \\
\midrule
4 & Task Oriented and Open Domain Dialogue & Research advances task-oriented and open-domain dialogue by combining end-to-end response generation with policy learning and state tracking, grounding on external knowledge and personas, handling complex goals, and developing robust evaluation. \\
\midrule
5 & LLM Mathematical and Logical Reasoning & Research advances methods and benchmarks to enhance and evaluate the mathematical and logical reasoning capabilities of large language models via chain-of-thought and structured prompting, neuro-symbolic and programming scaffolds, data/tool augmentation, self-correction, and reward modeling, and step-level and compositional evaluation. \\
\midrule
6 & Causal Discovery and Treatment Effects & Research develops methods to learn causal structure and estimate treatment and direct/indirect effects via interventions and counterfactuals, addressing hidden confounding, temporal dynamics, and identifiability in observational and experimental data. \\
\midrule
7 & Retrieval Augmented Question Answering & Research advances dense and generative retrieval and retrieval-augmented generation to enhance LLM question answering, addressing document identifiers, ranking and sampling, query rewriting and generation, knowledge selection, and rigorous evaluation for accurate and robust answers. \\
\midrule
8 & Advances in Neural Machine Translation & Research advances neural machine translation through improved architectures and decoding, multilingual and low-resource adaptation and transfer, robust evaluation and quality estimation, and human-in-the-loop and memory-assisted techniques. \\
\midrule
9 & Multimodal Visual Question Answering and Reasoning & Research advances vision-language models for multimodal VQA and reasoning, introducing datasets and benchmarks, alignment and retrieval strategies, and methods for compositional, chart, spatial, and knowledge-based tasks, and improving evaluation. \\
\midrule
10 & Text-to-Image Editing and Control & Research advances text-to-image diffusion with fine-grained controllability and personalization via improved architectures, guidance mechanisms, and language-vision integration, enabling precise editing, multi-concept synthesis, and preference-aligned generation. \\
\midrule
11 & Stochastic Bandits and Regret Bounds & Research develops algorithms and theoretical guarantees for stochastic, contextual, and linear bandits, including combinatorial, multi-agent, federated, and multi-objective variations, improving regret and sample complexity bounds under diverse feedback and constraints. \\
\midrule
12 & Explainable AI with Trees Concepts Counterfactuals & Research unifies decision tree modeling and optimization, prototype and concept-based self-explanation, Shapley/attribution and interaction metrics, and counterfactual and recourse frameworks to provide faithful, efficient, and user-relevant explainability. \\
\midrule
13 & Heterogeneity-Aware Personalized Federated Learning & This topic advances personalized FL by mitigating statistical and system heterogeneity through personalized models, heterogeneity-aware aggregation and alignment, and communication-efficient or asynchronous training, enhancing robustness, generalization, and scalability across clients and tasks. \\
\midrule
14 & LLMs for Code Intelligence & Research leverages LLMs with verification, search, and program analysis capabilities to generate, fix, translate, and retrieve code across languages, while benchmarking, evaluating robustness and bias, detecting hallucinations, and integrating developer feedback and tools. \\
\midrule
15 & Differential Privacy Algorithms and Utility Guarantees & Research advances differential privacy through novel mechanisms, optimization, and learning algorithms (e.g., adaptive clipping, shuffle amplification) to protect gradients and data in federated and centralized settings with formal guarantees and improved utility bounds. \\
\midrule
16 & Robotic Grasping and Manipulation Policies & Research develops learning-based policies for dexterous grasping and object manipulation, utilizing demonstrations, reinforcement learning, simulation, and generative models with multimodal sensing to achieve robust and generalizable control. \\
\midrule
17 & Temporal Action Recognition and Localization & Research on spatio-temporal modeling for recognizing and temporally localizing human actions in videos, emphasizing long-range dynamics, label-efficient learning (weak/semi/unsupervised), and robust representations against noise and backgrounds. \\
\midrule
18 & Agentic LLMs for Planning and Games & Research develops and evaluates LLM-based agents that plan, communicate, and act in games and interactive environments, integrating symbolic planning, multi-agent coordination, and experiential learning for robust decision-making. \\
\midrule
19 & Efficient Vision Transformers and Token Mixing & This topic advances efficient vision backbones by redesigning self-attention and token processing (via token pruning/merging, alternative mixers (linear/FFT), and state space models) to preserve global context while reducing computation and memory for classification, detection, and segmentation. \\
\bottomrule
\end{tabularx}
\label{tab: topic_detail}
\end{table*}

\section{Retrieval System Evaluation.}
This appendix details the validation of our $ResearchDB$ based retrieval system, focusing on the effectiveness of integrating multidimensional knowledge profiles compared to traditional baselines.

\noindent\textbf{Expert Rating and Scoring.}
We carefully selected 20 representative complex queries designed to utilize the multidimensional capabilities of $ResearchDB$. We engaged 5 PhD students and domain experts with over three years of research experience to serve as independent raters. Raters used a unified scoring template to evaluate the top 10 retrieval results for each query.
\begin{itemize}
    \item \textbf{Scoring Standard (0-5 Points):} Raters assigned a score from 0 to 5 based on the relevance of the retrieved paper to the query's multidimensional constraints:
    \begin{itemize}
        \item \textbf{5 (Perfectly Relevant):} Paper perfectly satisfies all specified dimensional constraints of the query.
        \item \textbf{4 (Highly Relevant):} Paper satisfies almost all constraints, with only a minor deviation.
        \item \textbf{3 (Moderately Relevant):} Paper satisfies the primary query intent but deviates in one or two secondary dimensions.
        \item \textbf{2 (Slightly Relevant):} Paper is only generally relevant by topic or keywords, but satisfies some minimal dimensional criteria.
        \item \textbf{1 (Minimally Relevant):} Paper is only generally relevant by topic or keywords, failing to meet most dimensional constraints.
        \item \textbf{0 (Not Relevant):} Paper is completely irrelevant to the query intent.
    \end{itemize}
\end{itemize}

\begin{figure}[htp]
    \centering
    
    \includegraphics[width=0.90\linewidth]{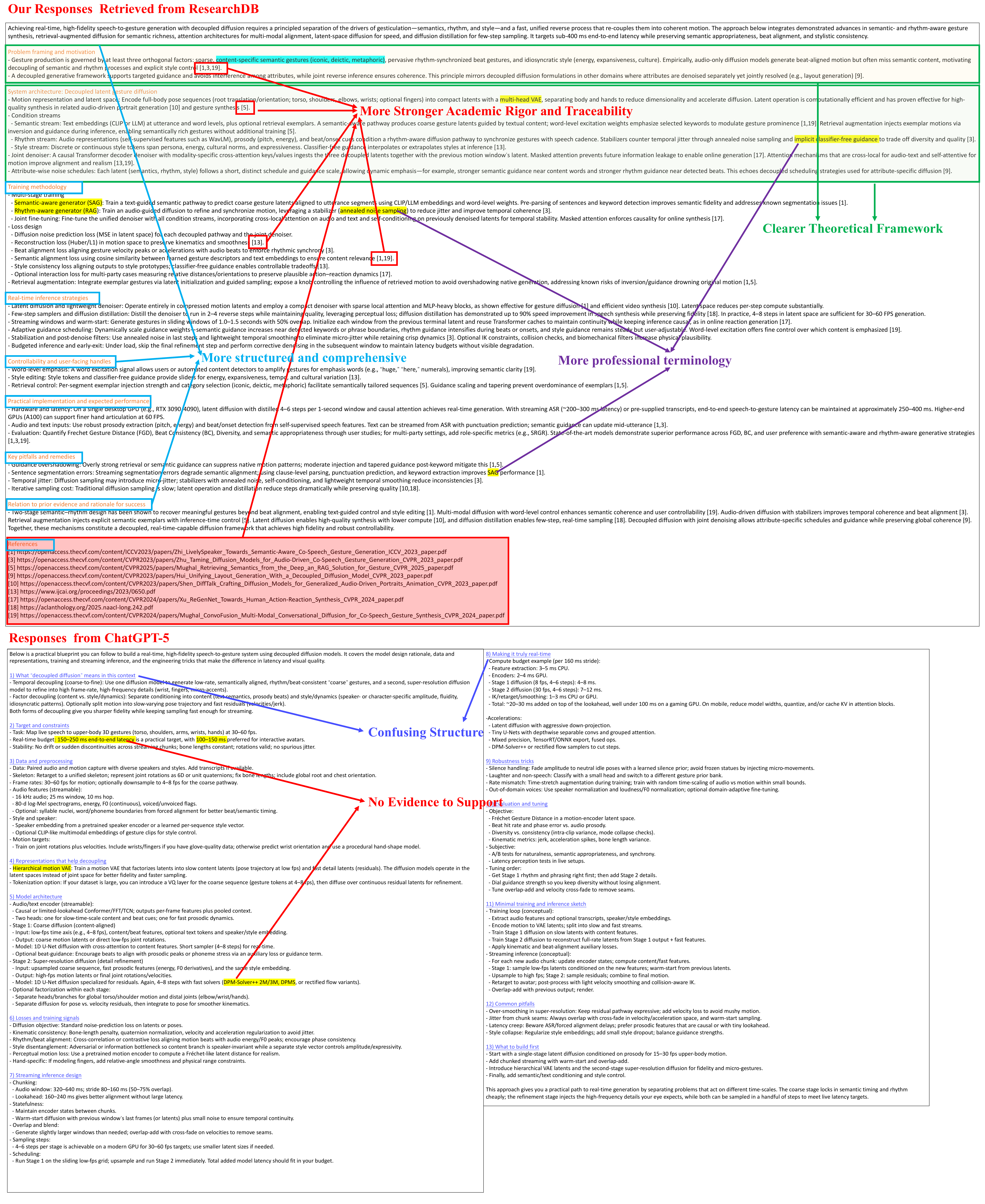}
    
    \caption{Qualitative comparison of answer quality between the $ResearchDB$ system and the ChatGPT-5 baseline on the query: ``How to achieve real-time, high-fidelity speech-to-gesture generation using decoupled diffusion models.'' }
    \label{fig:example}
    
\end{figure}

\noindent\textbf{Evaluation Results.}
Overall validation demonstrates that the $ResearchDB$ retrieval system, driven by multidimensional knowledge profiling, significantly outperforms traditional keyword or embedding-only retrieval baselines on complex, multi-faceted queries.
To illustrate this advantage, we conducted a qualitative assessment contrasting the $ResearchDB$ output with a ChatGPT-5 baseline on a complex query: ``How to achieve real-time, high-fidelity speech-to-gesture generation using decoupled diffusion models.'' The $ResearchDB$-supported answer (akin to an Academic Review) showed clear superiority in \cref{fig:example}:

\begin{enumerate}
    \item \textbf{Enhanced Academic Rigor and Traceability:} $ResearchDB$ successfully retrieved seminal papers, allowing the final answer to cite up to nine top-tier conference papers (e.g., [1], [3], [19]), accurately attributing key methods like Retrieval Augmentation.
    \item \textbf{Clearer Theoretical Framing:} The retrieved content enabled high-level theoretical decoupling of the problem (e.g., separating gesture drivers into semantic, rhythmic, and stylistic factors), resulting in a logically coherent survey argument.
\end{enumerate}
This qualitative evidence confirms that $ResearchDB$'s multidimensional matching enhances the output's Academic Robustness, ensuring the retrieved information is both relevant and authoritative for writing top-tier research reviews.

\noindent\textbf{Efficiency.}
To evaluate the scalability of ResearchDB, we benchmarked the knowledge profiling pipeline on a cluster equipped with 8 NVIDIA A100 (80GB) GPUs. 
\begin{itemize}
    \item \textbf{Knowledge Extraction Throughput:} By leveraging a distributed parallel processing framework with vLLM for inference acceleration, the system achieved a steady-state throughput of approximately 870 full papers per hour. The entire corpus of 104,482 papers was processed in 120 hours (5 days) of wall-clock time, totaling 960 GPU hours. 
    \item \textbf{Data Sharing and Accessibility:} To facilitate future research and ensure reproducibility, we will release the fully structured ResearchDB, containing multidimensional profiles (e.g., methods, datasets, and limitations) for all 104,482 papers. This pre-processed knowledge base allows researchers to perform large-scale trend analysis and complex semantic queries without incurring the high computational costs of initial parsing. 
    \item \textbf{Query Response:} For the end-user, the parallelized intent recognition module decomposes complex queries into sub-tasks that are executed concurrently, maintaining an average end-to-end response time of $\sim$1.2 seconds. 
\end{itemize}

\end{document}